\documentclass[10pt,twocolumn,letterpaper]{article}

\usepackage[pagenumbers]{cvpr} 

\definecolor{cvprblue}{rgb}{0.21,0.49,0.74}
\usepackage[pagebackref,breaklinks,colorlinks,allcolors=cvprblue]{hyperref}

\def\confName{CVPR}
\def\confYear{2027}

\usepackage{amsmath,amsfonts,bm}

\def\eqref#1{equation~\ref{#1}}

\def\1{\bm{1}}

\newcommand{\test}{\mathcal{D_{\mathrm{test}}}}

\DeclareMathAlphabet{\mathsfit}{\encodingdefault}{\sfdefault}{m}{sl}
\SetMathAlphabet{\mathsfit}{bold}{\encodingdefault}{\sfdefault}{bx}{n}

\usepackage{hyperref}
\usepackage{url}
\usepackage[dvipsnames]{xcolor}
\usepackage{amsmath} 
\usepackage{bbm} 
\usepackage{siunitx} 
\usepackage{kotex} 
\usepackage{array,multirow,graphicx}
\usepackage{float}
\usepackage{amsmath}
\usepackage{mathtools}
\usepackage{bbm}
\usepackage{hhline}
\usepackage{boldline}
\usepackage{tabularx}
\usepackage{float}
\usepackage{comment}
\usepackage{color}
\usepackage{amssymb}
\usepackage{multicol}
\usepackage{microtype}      
\usepackage{xcolor}         
\usepackage{algorithm}
\usepackage{algpseudocode}
\usepackage{pifont}
\usepackage{soul} 
\usepackage{subcaption}
\usepackage{wrapfig}
\usepackage{lipsum}
\usepackage{booktabs}
\usepackage{xcolor,colortbl}
\usepackage{nicematrix}
\usepackage{kotex}
\usepackage{dsfont}
\usepackage{caption}
\usepackage[accsupp]{axessibility}
\usepackage{amsmath}
\usepackage{subcaption}
\usepackage[table]{xcolor} 
\colorlet{oursrow}{red!20}
\newcommand{\hlcaption}[2]{%
    \setlength{\fboxsep}{1.5pt}
    \colorbox{#1}{#2}
}
\usepackage{xspace}            
\providecommand{\ours}{FactEmbed\xspace} 
\providecommand{\eg}{\emph{e.g.}\xspace}
\providecommand{\ie}{\emph{i.e.}\xspace}

\newcommand{\Lnce}{\mathcal{L}_{\mathrm{NCE}}}

\newcommand{\eos}{\texttt{[EOS]}\xspace}

\newcommand\blfootnote[1]{%
  \begingroup
  \renewcommand\thefootnote{}\footnote{#1}%
  \addtocounter{footnote}{-1}%
  \endgroup
}
\definecolor{bestred}{HTML}{FFC3C3}
\definecolor{secondorange}{HTML}{FFD9B2}
\definecolor{thirdyellow}{HTML}{FFFFB2}

\title{Factorized MLLM Embeddings for Composed Image Retrieval}

\author{WonJun Moon$^1$ \quad Seongchan Kim$^1$ \quad Dahun Kim$^2$ \quad Seungryong Kim$^{1, \ast}$ \\
KAIST AI$^1$ \quad Google DeepMind$^2$\\
\small
\url{https://cvlab-kaist.github.io/FactEmbed/}
}

\begin{document}

\maketitle

\begin{abstract}
Composed Image Retrieval~(CIR) aims to retrieve a target image given a reference image and a modification text. 
Although recent approaches leveraging Multimodal Large Language Models~(MLLMs) have shown significant progress, we argue that the MLLM's compositional understanding remains underutilized. 
This stems from directly mapping composed queries into a retrieval embedding via contrastive supervision alone, without explicitly resolving the intended visual modifications.
In this work, we introduce Factorized MLLM Embeddings~(\ours), which decouples compositional query comprehension from building discriminative embeddings.
Rather than directly projecting the composed query into a single embedding without internal reasoning, \ours employs a set of factorized latents that simulate the intended visual modifications via an intermediate target imagination objective. 
This process encourages the latents to ground into target semantics before forming discriminative representations for retrieval. 
Subsequently, to propagate these factorized representations throughout the model for retrieval, we optimize a factor-level contrastive loss while distilling the decomposed representations into a deeper attention layer of the MLLM.
Finally, for efficient inference, we retain the \eos token to establish a global-level embedding on top of these factorized latents, while we exploit the mean-pooled factorized latents to rerank a short candidate list at negligible cost, as both are produced by the same forward pass.
Extensive evaluations demonstrate that \ours outperforms prior works across CIR benchmarks.
\vspace{-15pt}
\blfootnote{
$^\ast$ Corresponding author.
}
\end{abstract}

\section{Introduction}
Composed Image Retrieval~(CIR) aims to retrieve a target image from a gallery using a composed query, consisting of a reference image and a modification text~\citep{TIRG}. 
CIR allows users to express their search intent through relative descriptions: they may provide an image and specify how the desired result should differ from it.
This paradigm is appealing when users can recognize a desired image by its visual relation to a reference, but cannot easily describe the target itself. 

Recently, Multimodal Large Language Models~(MLLMs) have emerged as a powerful foundation for CIR, owing to their ability to deeply fuse visual and linguistic information within a unified autoregressive representation space~\citep{RECALL, jiang2024e5}. 
A common strategy is to finetune an MLLM contrastively by compressing the multimodal query and the target image into a single embedding, typically obtained from the final \eos token, and directly optimizing their similarity~\citep{CIR-LVLM, liu2025lamra,zhang2024gme, RECALL}. 
Despite its effectiveness, this formulation exposes a key limitation: forcing a direct mapping onto a single embedding provides no explicit signal for the MLLM to exploit its compositional understanding capability to resolve the requested edit before forming a discriminative embedding.

To overcome this, we aim to decouple query comprehension from discriminative embedding learning, allowing the model to first simulate the requested edit, \ie, imagine the target, and then build its retrieval embedding upon the imagined target. We present \ours, which realizes this intermediate step through a compact set of factorized latents that parse the query before producing the final embedding.
These factorized latents are prepended to the \eos token in the input sequence, serving as information bottlenecks that decompose fused multimodal queries into factor-level components. 
This explicit decomposition enables \ours to isolate specific visual attributes and distinguish the factors the text modifies from those it preserves, in contrast to direct contrastive mapping which only implicitly infers the modified embedding.
To encourage each latent to capture different details in the target image, we adopt a target imagination objective parameterized by a lightweight decoder over these latents. 
Given a coarse layout prior of the target, the latents are trained to imagine the residual fine-grained factors that the prior cannot explain.
Constraining reconstruction through these bottlenecks forces the latents to cluster semantically coherent target factors to maximize the coverage.
This provides structural supervision without requiring explicit reasoning traces or dense annotations, thereby allowing the model to implicitly simulate the specified edit.

Reconstruction alone, however, shapes the latents generatively without ensuring their utility for retrieval: paired query and target could be faithfully decomposed yet embedded in incompatible coordinate spaces. 
We address this by adapting target imagination for retrieval in two ways. 
First, we apply a factor-level InfoNCE objective directly on the factorized latents, ensuring that the decomposed factors are aligned for retrieval rather than only for reconstruction.
Second, we distill the decoder’s factorized representation into the attention maps of the latents at a deeper layer, transferring the fine-grained decomposition from the auxiliary head into the backbone layers accessed during retrieval.

Lastly, we assign complementary roles to the two outputs produced in a single forward pass. 
Positioned prior to the \eos, the factorized latents allow \eos to integrate the decomposed factors into a global retrieval embedding, while \eos is optimized as a single retrieval embedding for efficient retrieval.
Yet, as fine-grained information is inevitably diluted when aggregating multiple factors into a single discriminative embedding, we rerank a short candidate list retrieved by \eos using the mean-pooled latents. 
This complementary structure allows \ours to achieve fast global recall via single \eos token while precision is significantly boosted by recovering fine-grained factor evidence during reranking.
Because both representations originate from a single forward pass and the mean-pooled latent adds only one vector per image, this design complements the global embedding with factor-level evidence without additional encoding or heavy storage overhead.
Combined with our MLLM data augmentation, \ours achieves state-of-the-art results on CIR benchmarks, outperforming both CIR methods and general-purpose embedding models.

\section{Related Work}
\noindent\textbf{Composed Image Retrieval.}
CIR aims to retrieve a target image from a gallery given a reference image with a modification text that states how the desired result should differ from it~\citep{TIRG, cirr, mahadev2026pinpoint, gu2023compodiff}. 
Earlier works~\citep{SPRC, TIRG, ARTEMIS, TGCIR, LIMN, CaLa} attempted to resolve representational conflict by performing the composition inside CLIP~\citep{CLIP} or BLIP~\citep{li2022blip, BLIP2} embeddings.
Then, since collecting high-precision annotations for CIR incurs significant cost~\citep{zhang2024magiclens, huynh2025collm}, two parallel lines of research have been actively explored: learning with noisy CIR annotations~\citep{INTENT} and zero-shot approaches~\citep{saito2023pic2word}.
While the former stream attempted to mitigate the impact of noisy annotations by leveraging loss dynamics~\citep{TME}, similarity curves~\citep{QuRe}, MLLM-constructed anchor set~\citep{Air-Know}, or unlearning process~\citep{ConeSep}, zero-shot approaches aim to bypass the high cost of collecting annotated triplets~\citep{jang2024slerp, baldrati2023searle, gu2024lincir, karthik2024visionbylanguage, imagineandseek, tang2025predicir}.
Notably, recent zero-shot methods also anticipate the target, either by generating a proxy image~\citep{imagineandseek} or by leveraging a world model~\citep{tang2025predicir}; in contrast, our imagination is embedded within the embedding model itself, requiring neither a generative model nor an additional inference pass.

More recently, LLMs and MLLMs have been widely adopted in CIR owing to their strong comprehension capabilities.
CIR-LVLM~\citep{CIR-LVLM} adapted a local MLLM for CIR by introducing intent-aware instructions with learnable soft prompts, while ReCALL~\citep{RECALL} proposed MLLM-based triplet augmentation strategy by diagnosing the blind spots of adapted MLLM retriever.
To further exploit the comprehension capability of LLMs, CCIN~\citep{CCIN} and FIRE-CIR~\citep{garderes2026fire} incorporated an external LLM at query time to decompose the query at the attribute level, which, however, incurs substantial inference overhead, with FIRE-CIR further requiring verification linear in the number of candidates.
Thus, in this work, we adapt a local MLLM to exploit its inherent comprehension capability; whereas prior MLLM-based methods~\citep{CIR-LVLM, RECALL} directly map the composed query onto a single embedding, we encourage the model to decompose the query within a single forward pass and build the retrieval embedding on top of the decomposed factors.

\noindent\textbf{Multimodal Retrieval with MLLMs.}
MLLM embeddings are being extensively studied in multimodal retrieval where complex comprehension is required~\citep{meng2025vlm2vecv2, zhou2025megapairs, qwen3vlembedding}.
VLM2Vec~\citep{jiang2025vlm2vec} initially used the last hidden state of the \eos token of a causal MLLM.
Yet, as \eos is not designed to generate embeddings, different methods are being studied, \textit{e.g.}, mean-pooling~\citep{lee2025gemini, vera2025embgemma, li2026umarvel} and multi-vector embeddings~\citep{liu2026crem, khattab2020colbert, faysse2025colpali}.
Also, integrating computation via explicit reasoning~\citep{hao2026trace}, latent rollouts~\citep{he2026plume}, or reinforcement-learned generative embeddings~\citep{lan2026umer1} is another prominent line of research.
Our work resembles the stream of appending a small bank of learned tokens whose hidden states become the representation~\citep{xiao2026metaembed, sun2026bottleneck}. 
None of these works, however, constrains what an individual token should represent.
In contrast, we leverage target reconstruction as a structural supervisory signal, forcing each factorized latent to ground and specialize in spatially coherent regions of the target scene.

\begin{figure*}
    \centering
    \vspace{-0.25cm}
    \includegraphics[width=1.\textwidth]{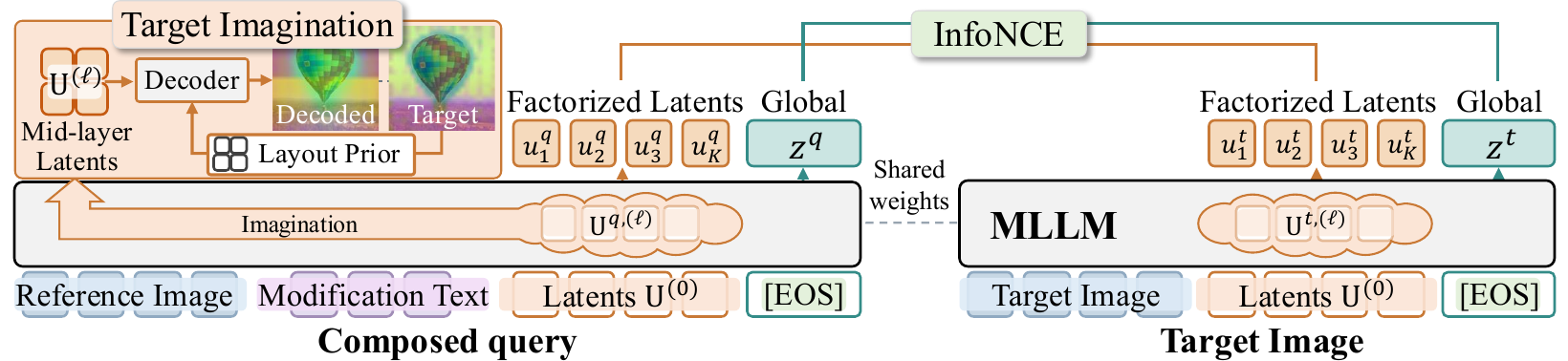}
    \vspace{-17pt}
    \caption{
    \textbf{Overview of \ours.}
    Given a composed query and a target image, \ours introduces a compact set of learnable factorized latents into the input sequence.
    To enforce structured factorization without requiring dense annotations, we attach a decoder to an intermediate MLLM layer, which utilizes these latents as an information bottleneck to reconstruct target patch features via a target imagination objective.
    In addition, we apply a factor-level InfoNCE to the resulting latents $\{u_k\}_{k=1}^{K}$ to leverage the decomposed factors for retrieval, while global-level embedding $z$ is trained with the instance-level InfoNCE objective.
    }
    \label{fig:main}
    \vspace{-13pt}
\end{figure*}

\noindent\textbf{Compact Latent Representations.}
Representing an input as a small set of latents has been studied extensively across fields.
Retrievers used it for late interaction~\citep{AleaVR, protoprvr, TMVM, ProST}, Perceiver~\citep{jaegle2021perceiver, jaegle2022perceiverio} exploited latents to decouple compute from input size, DETR~\citep{carion2020detr} used them as object queries to perform object detection, and slot attention~\citep{locatello2020slot, seitzer2023dinosaur, srl} utilized slots to decompose the scene into object-wise regions by normalizing the attention across latents.
\ours adopts the same competitive normalization in its decoder to induce factor-level specialization, but keeps the MLLM's forward pass intact, thereby preserving kernel optimizations such as FlashAttention~\citep{dao2024flashattention}.

\section{\ours: Factorized MLLM Embeddings}
\label{sec:method}

\noindent\textbf{Task.}
Given a query $q=(I_r, T)$ consisting of a reference image $I_r$ and a modification text $T$, the goal of Composed Image Retrieval~(CIR) is to retrieve the target image $t=I_t$ that depicts the reference after the described modification. 
The task is intrinsically compositional: the model must decide which visual factors of $I_r$ the text modifies~(\eg, ``replace the logo on the shirt'') and which it must preserve~(\eg, the rest of the shirt and the background), then express the result of that edit as a retrievable representation.


\noindent\textbf{Overview.}
As shown in Fig.~\ref{fig:main}, \ours adapts an MLLM to produce CIR embeddings, utilizing a shared backbone to encode both the multimodal query and the target image.
Rather than mapping the input directly to a single vector, we first introduce $K$ learnable factorized latents into the multimodal input, on top of which the \eos embedding is formed.
This provides the model with the capacity to represent the complex query as a structured set of compositional components.
To ground these latents in the visual content of the target image, we train them to reconstruct the target's patch features beyond a coarse layout prior, a process we refer to as target imagination.
We perform this reconstruction by attaching a lightweight decoder at an intermediate layer of the MLLM, which allows the deeper layers to transform the decomposed factors into discriminative representations for retrieval.
For retrieval training, we apply InfoNCE~\citep{simclr} at two levels: a factor-level loss encourages the factorized latents to contribute to retrieval by matching the query and target latents via optimal transport, while an instance-level loss optimizes the global embedding.
In this way, the imagined decomposition propagates fine-grained evidence into a compact embedding that remains efficient for retrieval.

\subsection{Factorized Latents with Target Imagination}
\label{sec:method:attributelatents}

\noindent\textbf{Motivation.}
CIR requires modifying specific visual concepts given text instructions~(\textit{e.g.}, replacing an item).
However, directly mapping a composed query to a global embedding in an MLLM provides no explicit mechanism for disentangling the visual attributes to be modified from those that should remain invariant.
To overcome this, we introduce a compact set of factorized latents whose explicit job is to parse the scene into structured factors.
By decomposing the representation, these latents are designed to give the model room to express the instructed modification on the relevant factors without distorting the others.

\noindent\textbf{Factorized Latents.}
Factorized latents are defined as a set of $K$ learnable tokens, $\mathbf{U}^{(0)} \in \mathbb{R}^{K \times D}$, which act as information bottlenecks to parse multimodal inputs into factorized representations. 
These tokens are prepended to the \eos in both the query and target sequences and interact with the multimodal inputs via the MLLM's causal attention.
We denote their hidden states at layer $\ell$ by $\mathbf{U}^{s,(\ell)} = [u^{s,(\ell)}_1, \dots, u^{s,(\ell)}_K]$ for $s \in \{q, t\}$, where $s$ indicates the query- or target-side input, and drop the layer index for the last layer, \ie, $\mathbf{U}^{s}$.

\noindent\textbf{Lightweight Decoder for Imagination.}
For target imagination, we employ a lightweight decoder parameterized by a self-attention layer and 3-layer MLP attached to an intermediate layer $\ell$~($\ell=14$ out of 28 layers by default) of the MLLM to reconstruct the target patch representation from a vision foundation model~\citep{oquab2023dinov2}. 
Note that we omit the superscript $s$ for brevity since the decoder is applied to each side independently.
Each 1D factorized latent $u^{(\ell)}_k \in \mathbb{R}^D$ is spatially broadcast across a 2D grid with learnable positional embeddings and passed through the decoder.
Yet, reconstructing the target from the latents alone would force them to encode the global layout of the target, which is dominated by low-frequency signals~(often irrelevant to the localized edits in CIR).
Thus, we condition the decoding process on a layout prior that provides only the coarse layout of the target; ground-truth~(GT) target features $F\in \mathbb{R}^{H \times W \times D_{\textrm{DINO}}}$ are average-pooled to $h \times h$, expanded back to full resolution, and added to the broadcast latents. 
As this layout prior already explains the coarse layout, minimizing the reconstruction error drives the latents to carry only the information that is not contained in the coarse layout.
Consequently, for each latent $k$, the decoder predicts spatial logits $\mathcal{M}_k\in\mathbb{R}^{H\times W}$ along with reconstruction features $\hat{F}_k\in\mathbb{R}^{H\times W\times D_{\text{DINO}}}$, where we index the $HW$ spatial positions by $p$.
To ensure that the contributions of all latents sum to 1 at each position $p$, we obtain the normalized alpha mask $\bar{\mathcal{M}}_k$ via a softmax over the latent dimension:
\begin{equation}
\bar{\mathcal{M}}_{k,p} = \frac{\exp(\mathcal{M}_{k,p})}{\sum_{k'=1}^{K}\exp(\mathcal{M}_{k',p})}.
\end{equation}
Then, the reconstructed feature is formed by the mask-weighted sum:
\begin{equation}
\hat{F} = \sum\nolimits_{k=1}^{K} \bar{\mathcal{M}}_k \odot \hat{F}_k,
\end{equation}
where $\odot$ is element-wise multiplication with broadcasting along the channel dimension.
Let $\hat{F}^{q}$ and $\hat{F}^{t}$ denote the features predicted from the query- and target-side latents, respectively. 
Both are supervised against the same GT target features $F$, and the imagination loss is the mean squared error averaged over both sides:
\begin{equation}
\label{eq:recon}
\mathcal{L}_{\mathrm{imag}} = \frac{1}{2} 
\sum\nolimits_{s\in\{q,t\}} \text{MSE}(\hat{F}^{s}, F).
\end{equation}
As the coarse target layout prior is shared by both sides, $\mathcal{L}_{\mathrm{imag}}$ is minimized only by improving the residual details that are not covered by the layout prior. 
On the query side, the latents must infer this residual from $(I_r, T)$ without observing the target, which is the component we refer to as imagination. 
On the target side, the same objective anchors the target latents to the same residual factors, so that query and target latents occupy comparable factor slots for the latent matching in Sec.~\ref{sec:method:ImagforRet}.

\subsection{From Imagination to Retrieval}
\label{sec:method:ImagforRet}
\noindent\textbf{Factor-level Matching.}
To explicitly involve the factorized latents in retrieval, we apply an InfoNCE~\citep{simclr} on top of them.
Building upon the factor alignment established by $\mathcal{L}_{\mathrm{imag}}$, we define the matching score  $s_{\mathrm{OT}}(\mathbf{U}^q,\mathbf{U}^t)$ using Sinkhorn optimal transport~(OT) over pairwise cosine similarities to allow flexible factor-to-factor correspondences~(see Appendix E).
For the $i$-th query and $j$-th target in a batch of size $B$, the factor-level similarity is $\mathbf{S}^{f}_{ij} = s_{\mathrm{OT}}(\mathbf{U}^q_i, \mathbf{U}^t_j)$.
With the batch-wise contrastive loss over a similarity matrix $\mathbf{S}\in\mathbb{R}^{B\times B}$ defined as:
\begin{equation}
  \mathcal{C}(\mathbf{S}) = -\frac{1}{B}\sum_{i=1}^{B}
  \log \frac{\exp(\mathbf{S}_{ii}/\tau)}{\sum_{j=1}^{B}\exp(\mathbf{S}_{ij}/\tau)},
\label{eq:nce_generic}
\end{equation}
applying $\mathcal{C}(\mathbf{S}^{f})$ encourages each query factor to align with its counterpart in the target, which a single pooled vector cannot enforce.
$\tau$ is a temperature parameter.

\noindent\textbf{Instance-level Matching.}
The \eos token, which attends to the factorized latents, aggregates the comprehended factors into a single embedding.
We read out its normalized hidden state as the global embedding $z$ and apply the same contrastive loss $\mathcal{C}(\mathbf{S}^{g})$ with the instance-level similarity $\mathbf{S}^{g}_{ij} = z_i^{q\top} z_j^{t}$ between the $i$-th query and the $j$-th target. 
Unlike the factorized set, which requires storing $K$ vectors per target image, $z$ remains a single vector and thus serves as the primary retrieval embedding at scale.
Our retrieval objective thus combines both levels:
\begin{equation}
  \Lnce = \mathcal{C}(\mathbf{S}^{f}) + \mathcal{C}(\mathbf{S}^{g}),
\label{eq:nce}
\end{equation}
where $\mathcal{C}(\mathbf{S}^{f})$ aligns individual factors through OT and $\mathcal{C}(\mathbf{S}^{g})$ directly optimizes the embedding used at test time.


\noindent\textbf{Propagating Factorization into the MLLM Backbone.}
While intermediate-layer decoding effectively encourages latent specialization, deeper layers near the retrieval output tend to lose spatial decomposition as representations abstract into global embeddings. 
To preserve factorized knowledge within the MLLM backbone, we distill the spatial alpha masks $\bar{\mathcal{M}}$ generated by the decoder into the latent-to-patch attention of the factorized latents at a deeper layer. 
Note that we formulate the spatial distillation loss exclusively on the target-side forward pass, as the query-side attention maps are defined over the reference image space, which does not share a spatial grid with the target alpha masks.
Specifically, let $\mathcal{A}_k \in \mathbb{R}^{H \times W}$ denote the head-averaged attention of the $k$-th factorized latent toward the target patch tokens at layer $\ell_{\text{top}}$, with the softmax taken over the patch keys only, bilinearly resampled to the decoder grid. 
Then, we normalize $\mathcal{A}_k$ and $\bar{\mathcal{M}}_k$ over the spatial positions into distributions $\mathbf{a}_k$ and $\mathbf{m}_k$, each describing the region the $k$-th latent occupies:
\begin{equation}
\mathbf{a}_{k,p}=\frac{\mathcal{A}_{k,p}}{\sum_{p'}\mathcal{A}_{k,p'}},\quad
\mathbf{m}_{k,p}=\frac{\bar{\mathcal{M}}_{k,p}}{\sum_{p'}\bar{\mathcal{M}}_{k,p'}}.
\end{equation}
Consequently, we formulate a spatial distillation loss using Kullback-Leibler~(KL) divergence to align the deeper layer attention distributions with the decoder's spatial alpha masks:
\begin{equation}
\mathcal{L}_{\mathrm{distill}} = \frac{1}{K}\sum_{k=1}^{K}\mathrm{KL}\big(\mathbf{m}_k \,\|\, \mathbf{a}_k\big).
\label{eq:attn}
\end{equation}
This explicitly transfers spatial decomposition from the decoder to the MLLM.
Although the supervision is applied only on the target side, it shapes the backbone and the latents shared by both sides, thereby encouraging the query-side latents to also retain localized semantics at inference, where the target layout prior is unavailable.

\begin{table*}[t]
    \centering
    \vspace{-6pt}
    \caption{\textbf{Results on Refined-CIRR and Refined-FashionIQ.} 
    $\dagger$ are the results from CoLLM~\citep{huynh2025collm}.
    All other baselines are reproduced using official checkpoints, except for QWEN2.5-VL and ReCALL, which were re-run from scratch due to the lack of available checkpoints.
    $\ast$ denotes the use of additional gallery image captions.
    ``Avg.'' for Refined-CIRR is computed as $(R@1 + R_{\mathrm{subset}}@1)/2$, while on Refined-FashionIQ, it represents the mean across all three categories.
    Best and second-best results are shown in \textbf{bold} and \underline{underlined}, respectively.
    }
    \label{tab:refined_results_combined}
    \setlength{\aboverulesep}{0.2ex}  
    \setlength{\belowrulesep}{0.2ex}  
    \newcommand\headstrut{\rule[-1.1ex]{0pt}{3.5ex}}  
    \scriptsize
    \vspace{-4pt}
    \renewcommand{\arraystretch}{1.}  
    \setlength{\tabcolsep}{4.6pt} 
    \begin{tabular}{l cccc ccc c cc cc cc cc}
        \toprule
        \multirow{3}{*}{Method}
        & \multicolumn{8}{c}{Refined-CIRR}
        & \multicolumn{8}{c}{Refined-FashionIQ} \\
        \cmidrule(lr){2-9} \cmidrule(lr){10-17}
        & \multicolumn{4}{c}{Recall@N}
        & \multicolumn{3}{c}{Recall$_{\mathrm{subset}}$@N}
        & \multirow{2}{*}{Avg.}
        & \multicolumn{2}{c}{Dress}
        & \multicolumn{2}{c}{Shirt}
        & \multicolumn{2}{c}{Top\&Tee}
        & \multicolumn{2}{c}{Avg.} \\
        \cmidrule(lr){2-5} \cmidrule(lr){6-8} \cmidrule(lr){10-11} \cmidrule(lr){12-13} \cmidrule(lr){14-15} \cmidrule(lr){16-17}
        & R@1 & R@5 & R@10 & R@50
        & R@1 & R@2 & R@3 & 
        & R@10 & R@50
        & R@10 & R@50
        & R@10 & R@50
        & R@10 & R@50 \\
        \midrule 
        CoVR-2$^\dagger$ & 56.1 & -- & 91.1 & 97.9 & -- & -- & -- & -- & -- & -- & -- & -- & -- & -- & 54.2 & 72.9 \\
        CoLLM$^\dagger$ & 60.4 & -- & 92.6 & 98.2 & -- & -- & -- & -- & -- & -- & -- & -- & -- & -- & 57.2 & 76.4 \\
        VLM2Vec 7B & 53.39 & 83.50 & 90.96 & 97.89 & 78.16 & 91.65 & 96.72 & 65.78 & 18.59 & 37.41 & 30.57 & 50.21 & 26.98 & 47.61 & 25.38 & 45.08 \\
        VLM2Vec-v2 2B & 59.93 & 86.79 & 92.57 & 98.32 & 81.47 & 93.30 & 97.73 & 70.70 & 35.57 & 60.19 & 53.83 & 71.53 & 49.13 & 70.09 & 46.18 & 67.27 \\
        LIMN & 52.32 & 82.74 & 90.40 & 98.04 & 74.41 & 90.55 & 96.28 & 63.37 & 53.27 & 74.43 & 60.03 & 78.73 & 63.84 & 82.36 & 59.05 & 78.50 \\
        ENCODER & 57.76 & 86.15 & 92.80 & 98.75 & 82.76 & 94.53 & 97.94 & 70.26 & 56.62 & 78.50 & 62.76 & 80.78 & 66.07 & 84.64 & 61.82 & 81.31 \\
        TME & 65.05 & 90.53 & 94.76 & 98.91 & 86.86 & 95.93 & 98.50 & 75.96 & 63.15 & 82.19 & 70.48 & 86.03 & 72.58 & 88.38 & 68.74 & 85.53 \\
        MM-Embed 8B & 60.64 & 87.73 & 93.66 & 98.57 & 81.54 & 92.64 & 97.35 & 71.09 & 52.21 & 74.37 & 63.71 & 81.67 & 64.98 & 83.88 & 60.30 & 79.97 \\
        Qwen3-VL-Emb 8B & 67.03 & 90.45 & 95.37 & 98.93 & 84.60 & 94.76 & 98.12 & 75.82 & 63.71 & 80.96 & 69.75 & 85.29 & 72.37 & 86.48 & 68.61 & 84.25 \\
        Core-Emb 8B & 64.00 & 88.85 & 93.30 & 98.73 & 81.11 & 93.05 & 97.53 & 72.56 & 59.13 & 78.11 & 66.54 & 81.93 & 66.34 & 83.06 & 64.00 & 81.04 \\
        ReMatch 3B & 62.40 & 89.36 & 94.04 & \underline{98.98} & 83.17 & 94.35 & 98.17 & 72.79 & 33.50 & 57.40 & 50.53 & 69.43 & 46.85 & 66.18 & 43.63 & 64.34 \\
        CSMCIR$^\ast$ & 67.16 & 90.96 & \textbf{95.44} & \textbf{99.06} & 86.15 & 95.75 & \underline{98.55} & 76.66 & 63.65 & 82.08 & \underline{71.48} & \underline{86.61} & \underline{73.51} & \textbf{88.82} & \underline{69.55} & \underline{85.83} \\
        \hline
        Qwen2.5-VL~(Base) & 64.64 & 89.51 & 94.53 & 98.91 & 84.65 & 94.68 & 97.86 & 74.65 & 61.14 & 80.74 & 69.01 & 84.40 & 70.30 & 85.50 & 66.82 & 83.55 \\
        ReCALL & \underline{67.41} & \underline{90.99} & 95.24 & \textbf{99.06} & \underline{87.12} & \underline{96.00} & \underline{98.55} & \underline{77.27} & \underline{64.21} & \underline{82.47} & 71.17 & 86.08 & 73.07 & \underline{88.55} & 69.48 & 85.70 \\
        \rowcolor{red!20}
        \textbf{FactEmbed~(Ours)} & \textbf{68.99} & \textbf{91.29} & \underline{95.42} & \textbf{99.06} & \textbf{87.45} & \textbf{96.16} & \textbf{98.63} & \textbf{78.22} & 
        \textbf{66.61} & \textbf{83.08} & \textbf{72.79} & \textbf{87.66} & \textbf{74.00} & 88.49 & \textbf{71.13} & \textbf{86.41} \\
        \bottomrule
    \end{tabular}
    \vspace{-9pt}
\end{table*}

\subsection{Training Objective and Inference}
\label{sec:method:objective}
\textbf{Training.}
Our joint objective reconciles two competing dynamics: $\mathcal{L}_{\mathrm{imag}}$ forces factorized latents to partition target features into disjoint factors, whereas $\mathcal{L}_{\mathrm{NCE}}$ encourages latents within an instance to collapse into a single discriminative embedding~\citep{moon2026mitigating}. 
To address this, we employ a staged learning strategy.
In Stage 1, we optimize only the reconstruction loss $\mathcal{L}_{\mathrm{imag}}$ to warm up the factorized latents and the decoder, establishing a stable factorized prior by mapping latents to a spatially coherent region of the target scene.
In Stage 2, we further train the MLLM backbone with DoRA~\cite{liu2024dora} adapters, jointly with the latents and the decoder, using full objective:
\begin{equation}
\mathcal{L}_{\mathrm{total}} = \mathcal{L}_{\mathrm{NCE}} + \lambda_{\mathrm{imag}}\mathcal{L}_{\mathrm{imag}} + \lambda_{\mathrm{distill}}\mathcal{L}_{\mathrm{distill}},
\label{eq:total}
\end{equation}
where $\lambda_{\mathrm{imag}}$ and $\lambda_{\mathrm{distill}}$ are coefficients. 
Finally, we follow ReCALL~\citep{RECALL} to add a post-training stage which integrates MLLM-augmented triplets to reduce false negatives~\citep{garderes2026fire, QuRe} and mitigate the influence of mislabeled triplets~\citep{TME, ConeSep, INTENT}, both of which stem from the sparse supervision of CIR datasets.
In brief, we adopt the existing augmentation pipeline to generate modification texts between each reference image and its mined hard negatives; the difference from \citep{RECALL} is that we condition the MLLM on the original target image and prompt it to describe what distinguishes the hard negative, which yields captions non-redundant with the original modification text in a single generation pass.
Details on the augmentation are in Appendix F.

\noindent\textbf{Inference.}
During inference, a single forward pass yields both global embedding $z$ and factorized latents $\{u_k\}_{k=1}^{K}$.
To minimize the memory footprint and computational cost, we aggregate latents into a single mean vector $\tilde{u} = \frac{1}{K}\sum_{k=1}^{K} u_k$, and utilize it exclusively for efficient reranking.
Note that $\tilde{u}$ is not explicitly optimized during training; rather, it serves as a compact proxy for the factorized set, whose members are aligned via OT, and thus requires no additional training.
Specifically, candidates in the gallery are initially searched with $z$ alone, and the top-100 candidate shortlist is reranked by integrating the cosine similarity of the mean factor as:
\begin{equation}
r(q,t) = \cos\big({z}^q, {z}^t\big) + \cos\big(\tilde{u}^q, \tilde{u}^t\big).
\label{eq:inference_short}
\end{equation}
Since only the top candidates are rescored with a single vector, the additional cost is negligible and independent of the gallery size, while still leveraging the factor-level evidence.





\begin{table*}[t]
    \centering
    \setlength{\aboverulesep}{0.2ex}  
    \setlength{\belowrulesep}{0.2ex}  
    \newcommand\headstrut{\rule[-1.1ex]{0pt}{3.5ex}}  
    \renewcommand{\arraystretch}{1.00}  
    \setlength{\tabcolsep}{2.5pt} 
    \scriptsize
    \vspace{-8pt}
    \begin{minipage}[t]{0.465\textwidth}
        \centering
        \vspace{-1pt}
        \caption{\textbf{Results on CIRR test set.}
        ``Avg.'' for CIRR test set is computed as $(R@5 + R_{\mathrm{subset}}@1)/2$.
        }
        \label{tab:cirr}
        \vspace{-6pt}
        \scriptsize
    \begin{tabular}{l cc cc cc cc}
        \toprule
        \multirow{2}{*}{Method}
        & \multicolumn{4}{c}{Recall@N}
        & \multicolumn{3}{c}{Recall$_{\mathrm{subset}}$@N}
        & \multirow{2}{*}{Avg.} \\
        \cmidrule(lr){2-5}
        \cmidrule(lr){6-8}
        & R@1 & R@5 & R@10 & R@50
        & R@1 & R@2 & R@3 & \\
        \midrule

        ARTEMIS~\citep{ARTEMIS}
        & 16.96 & 46.10 & 61.31 & 87.73
        & 39.99 & 62.20 & 75.67 & 43.05 \\

        TG-CIR~\citep{TGCIR}
        & 45.25 & 78.29 & 87.16 & 97.30
        & 72.84 & 89.25 & 95.13 & 75.57 \\

        SPRC~\citep{SPRC}
        & 51.96 & 82.12 & 89.74 & 97.69
        & 80.65 & 92.31 & 96.60 & 81.39 \\

        LIMN~\citep{LIMN}
        & 43.64 & 75.37 & 85.42 & 97.04
        & 69.01 & 86.22 & 94.19 & 72.19 \\

        CoVR-2~\citep{CoVR-2}
        & 50.43 & 81.08 & 88.89 & 98.05
        & 76.75 & 90.34 & 95.78 & 78.92 \\

        CaLa~\citep{CaLa}
        & 49.11 & 81.21 & 89.59 & 98.00
        & 76.27 & 91.04 & 96.46 & 78.74 \\

        ENCODER~\citep{ENCODER}
        & 46.10 & 77.98 & 87.16 & 97.64
        & 76.92 & 90.41 & 95.95 & 77.45 \\

        CIR-LVLM~\citep{CIR-LVLM}
        & 53.64 & 83.76 & 90.60 & 97.93
        & 79.12 & 92.33 & 96.67 & 81.44 \\

        QuRe~\citep{QuRe}
        & 52.22 & 82.53 & 90.31 & 98.17
        & 78.51 & 91.28 & 96.48 & 80.52 \\

        CCIN~\citep{CCIN}
        & 53.41 & 84.05 & 91.17 & 98.00
        & -- & -- & -- & -- \\

        TME~\citep{TME}
        & 53.42 & 82.99 & 90.24 & 98.15
        & 81.04 & 92.58 & 96.94 & 82.01 \\
        
        Air-Know~\citep{Air-Know}
        & 51.97 & 82.25 & 90.42 & 98.25 & 79.21 & 91.94 & 96.87 & 80.73 \\
        ConeSep~\citep{ConeSep}
        & 53.06 & 83.45 & 90.60 & 98.34 & \underline{81.23} & 92.62 & 97.24 & 82.34 \\

        Q3-VL-Emb 8B~\citep{qwen3vlembedding}$^\ddagger$
        & 51.95 & 80.92 & 88.58 & 97.49 & 75.23 & 90.29 & 95.66 & 78.08 \\ 

        CSMCIR$^\ast$~\citep{csmcir}
        & \underline{53.88} & \textbf{85.06} & \underline{91.69} & \textbf{98.58} & 80.10 & 92.24 & 96.92 & \underline{82.58} \\

        \hline 
        

        Q2.5-VL~(Base)~\citep{bai2025qwen25vl}$^\ddagger$ 
        & 51.98 & 81.90 & 90.27 & 98.19 & 77.69 & 91.85 & 96.92 & 79.80 \\

        ReCALL~\citep{RECALL} $^\ddagger$ 
        & 53.76 & 83.76 & 91.54 & 98.34 & 81.11 & \underline{93.35} & \underline{97.52} & 82.44 \\

        \rowcolor{red!20}
        \textbf{FactEmbed~(Ours)} & \textbf{56.60} & \underline{84.87} & \textbf{91.78} & \underline{98.41} & \textbf{81.86} & \textbf{93.78} & \textbf{97.61} & \textbf{83.37} \\

        \bottomrule
    \end{tabular}
    \end{minipage}
    \hfill
    \begin{minipage}[t]{0.515\textwidth}
        \centering
        \vspace{-1pt}
        \caption{\textbf{Results on FashionIQ validation set.}
        ``Avg.'' represents the mean performance across all three categories.
        }
        \label{tab:fiq}
        \vspace{-6pt}
        \scriptsize
        \begin{tabular}{lcccccccc}
        \toprule
        \multirow{2}{*}{Method}
        & \multicolumn{2}{c}{Dress}
        & \multicolumn{2}{c}{Shirt}
        & \multicolumn{2}{c}{Top\&Tee}
        & \multicolumn{2}{c}{Avg.} \\
        \cmidrule(lr){2-3}
        \cmidrule(lr){4-5}
        \cmidrule(lr){6-7}
        \cmidrule(lr){8-9}
        & R@10 & R@50
        & R@10 & R@50
        & R@10 & R@50
        & R@10 & R@50 \\
        \midrule



        FashionSAP~\citep{FashionSAP}
        & 33.71 & 60.43
        & 41.91 & 70.93
        & 33.17 & 61.33
        & 36.26 & 64.23 \\

        FAME-ViL~\citep{FAME-ViL}
        & 42.19 & 67.38
        & 47.64 & 68.79
        & 50.69 & 73.07
        & 46.84 & 69.75 \\

        SyncMask~\citep{SyncMask}
        & 33.76 & 61.23
        & 35.82 & 62.12
        & 44.82 & 72.06
        & 38.13 & 65.14 \\

        SADN~\citep{SADN}
        & 40.01 & 65.10
        & 43.67 & 66.05
        & 48.04 & 70.93
        & 43.91 & 67.36 \\

        CaLa~\citep{CaLa}
        & 42.38 & 66.08
        & 46.76 & 68.16
        & 50.93 & 73.42
        & 46.69 & 69.22 \\

        CoVR-2~\citep{CoVR-2}
        & 46.53 & 69.60
        & 51.23 & 70.64
        & 52.14 & 73.27
        & 49.96 & 71.17 \\

        SPRC~\citep{SPRC}
        & 49.18 & 72.43
        & 55.64 & 73.89
        & 59.35 & 78.58
        & 54.72 & 74.97 \\

        CIR-LVLM~\citep{CIR-LVLM}
        & 50.42 & 73.60
        & \textbf{58.59} & \underline{75.86}
        & 59.61 & 78.99
        & 56.21 & 76.14 \\

        CCIN~\citep{CCIN}
        & 49.38 & 72.58
        & 55.93 & 74.14
        & 57.93 & 77.56
        & 54.41 & 74.76 \\

        TME~\citep{TME}
        & 49.73 & 71.69
        & 56.43 & 74.44
        & 59.31 & 78.94
        & 55.15 & 75.02 \\

        QuRe~\citep{QuRe}
        & 46.80 & 69.81
        & 53.53 & 72.87
        & 57.47 & 77.77
        & 52.60 & 73.48 \\

        Air-Know~\citep{Air-Know}
        & 51.03 & 73.08 & 56.76 & 75.05 & 59.47 & 79.02 & 55.75 & 75.72 \\

        ConeSep~\citep{ConeSep}
        & 50.96 & 73.02 & 56.98 & 75.22 & 58.80 & 79.40 & 55.58 & 75.88 \\
        
        Q3-VL-Emb 8B~\citep{qwen3vlembedding}$^\ddagger$
        & 47.84 & 69.46 & 52.36 & 71.34 & 55.58 & 74.40 & 51.93 & 71.73 \\ 

        CSMCIR$^\ast$~\citep{csmcir}
        & \underline{52.45} & \textbf{74.81} & 57.70 & 75.76 & \underline{61.14} & \textbf{80.98} & \underline{57.10} & \textbf{77.18} \\

        \hline
        Q2.5-VL~(Base)~\citep{bai2025qwen25vl}$^\ddagger$ & 45.81 & 70.65 & 55.50 & 73.45 & 57.62 & 76.95 & 52.98 & 73.68 \\
        ReCALL~\citep{RECALL} $^\ddagger$ 
        & 48.93 & 72.63 & 54.76 & 73.60 & 57.98 & 78.74 & 53.89 & 74.99 \\
        \rowcolor{red!20}
        \textbf{FactEmbed~(Ours)} & \textbf{53.25} & \underline{74.67} & \underline{58.34} & \textbf{76.30} & \textbf{61.55} & \underline{79.65} & \textbf{57.71} & \underline{76.87} \\
        \bottomrule
    \end{tabular}
    \end{minipage}
    \vspace{3pt}
    \begin{minipage}{\textwidth}
        \footnotesize
        \textbf{Bold} and \underline{underline} denote the best and second-best results, respectively.
        $^\ddagger$ denotes results reproduced by us with official implementation.
        $^\ast$ denotes the additional use of image captions for gallery images, and Q denotes QWEN.
    \end{minipage}
    \vspace{-13pt}
\end{table*}

\section{Experiments}


\textbf{Datasets \& Metrics.}
We test \ours on CIRR~\citep{cirr} and FashionIQ~\citep{fashioniq}, two standard CIR benchmarks.
However, as original benchmarks contain ambiguous samples~\citep{huynh2025collm} and fail to provide robust evaluation~\citep{liu2026sanitycheckcomposedimage}, we conduct our primary evaluations on Refined-CIRR and Refined-FashionIQ~\citep{huynh2025collm}.
Results on original datasets are also reported.
For evaluation metrics, we use Recall@$N$~(R@$N$), the fraction of queries whose target appears in the top-$N$ results. 
Along with standard Recall, we also report Recall$_{\mathrm{subset}}@\{1,2,3\}$ on CIRR, which are computed within each query's curated six-candidate subset.

\noindent\textbf{Implementation Details.}
Adopting the setup from ReCALL's repository~\citep{RECALL}, we use DoRA~\citep{liu2024dora} to finetune Qwen2.5-VL-7B-Instruct~\citep{bai2025qwen25vl} with rank $16$, scaling factor $64$. 
We use AdamW optimizer, a cosine schedule with $10\%$ warmup, a batch size of 512, and the last checkpoint is used for evaluation.
For hyperparameters, we set the number of latents to $K=8$, the resolution of the target layout prior to $2\!\times\!2$~($h=2$), and the decoding layer to $\ell=14$. 
We set $\ell_{\text{top}}=-3$, \textit{i.e.,} the factorized map is distilled into the attention map of the third layer from the top.
Loss coefficients are set to $\lambda_{\mathrm{imag}}\!=\!1.0$ and $\lambda_{\mathrm{distill}}\!=\!0.5$ for all experiments.


\subsection{Main Results}
In Tab.~\ref{tab:refined_results_combined}, we benchmark \ours on refined CIR benchmarks against recent CIR methods and general-purpose embedding models~\citep{jiang2025vlm2vec, meng2025vlm2vecv2, mmembed, qwen3vlembedding, coreembed, rematch} using their official checkpoints.
While prior works primarily relied on broader metrics like Recall@5 on CIRR due to sample ambiguities in original benchmarks, we emphasize that Recall@1 and Subset Recall@1 on CIRR serve as more precise ranking metrics under these refined evaluation settings, where broader metrics saturate near top performance.
Specifically, \ours outperforms two strong baselines, CSMCIR~\citep{csmcir}~(which additionally relies on synthetic text captions for every gallery image) and ReCALL~\citep{RECALL}: it achieves absolute gains of 1.83 and 1.58 points in Recall@1 on Refined-CIRR, and 1.58 and 1.65 points in average Recall@10 on Refined-FashionIQ, respectively.
These gains demonstrate the advantage of decoupling complex query understanding from direct visual alignment: our decomposed factorized latents yield fine-grained, disambiguated semantic representations that standard contrastive architectures do not explicitly encourage.

In Tab.~\ref{tab:cirr} and Tab.~\ref{tab:fiq}, we further report performance on the original CIRR and FashionIQ datasets.
Notably, all methods score substantially lower on the original benchmarks than on their refined counterparts.
Excluding Qwen3-VL-Embedding 8B, this gap reaches at most 13.7 points in R@1 on CIRR and 15.6 points in average R@10 on FashionIQ, whereas Qwen3-VL-Embedding 8B, the state-of-the-art generalist model, exhibits the largest gap~(15.1 and 16.7 points, respectively).
This is consistent with prior findings~\citep{huynh2025collm, liu2026sanitycheckcomposedimage} that the original benchmarks contain ambiguous annotations and evaluation artifacts, which may underestimate the compositional reasoning capabilities of high-capacity models.
Nevertheless, \ours achieves the best overall result.

\subsection{Forming Embeddings with Factorized Latents}
\label{sec:ablation_fullattribute}
As factorized latents can be turned into a retrieval score in several ways, we compare the variants in Tab.~\ref{tab:attr_pooling}.
Matching the full latent set via OT~(a) achieves the highest R@1 on both benchmarks, but requires storing $K$ vectors per gallery image, whose storage and search costs grow with the gallery size~\citep{khattab2020colbert, faysse2025colpali}.
The single-vector alternatives are (b)--(d): (b) and (d) use only the mean factor $\tilde{u}$ and the global embedding $z$ read out from \eos of our final model, respectively, while (c) is a variant trained with the instance-level objective applied to $\tilde{u}$ in place of $z$.
To illustrate, we find that the mean factor~(b) performs worst, as $\tilde{u}$ is never directly optimized for retrieval.
Optimizing it directly~(c) improves over (b) but does not surpass $z$~(d), indicating that replacing \eos with instance-level supervised latents brings no benefit as a global embedding.
We therefore retain \eos for the global embedding and keep the latents factor-aligned for efficient reranking~(e), as the factor-aligned $\tilde{u}$ complements the instance-level global embedding $z$.
This efficient reranking narrows the R@1 gap to full factor matching to 0.24 and 0.21 on CIRR and Refined-CIRR, respectively, with only two vectors per gallery image, a memory footprint on par with standard multi-vector CIR methods~\citep{ConeSep, TME, CCIN, QuRe, SPRC}.
\begin{table}
    \centering
    \vspace{-10pt}
    \caption{
    \textbf{Inference strategy for balancing performance and efficiency.}
    While matching the full set of factorized latents via optimal transport yields the best performance, its memory footprint scales with the number of latents.
    Efficient reranking achieves comparable results to full factor matching while maintaining low overhead.
    ``Mean Factor$^\#$'' is a variant where the instance-level contrastive loss is applied to the mean-pooled latent $\tilde{u}$ instead of \eos, together with the factor-level loss.
    \protect\hlcaption{oursrow}{Red} highlights our default configuration.
    }
    \label{tab:attr_pooling}
    \vspace{-7pt}
    \scriptsize
\setlength{\aboverulesep}{0pt}
\setlength{\belowrulesep}{0pt}
\newcommand\headstrut{\rule[-1.1ex]{0pt}{3.5ex}}  
\newcommand\tstrut{\rule{0pt}{2.6ex}}             
\newcommand\bstrut{\rule[-0.9ex]{0pt}{0pt}}       
\renewcommand{\arraystretch}{0.95}
    \setlength{\tabcolsep}{5.1pt} 
    \scriptsize
        \begin{tabular}{c|l| cccccc}
            \toprule
            \headstrut & \multirow{2}{*}[-0.4ex]{Retrieval Unit} & \multicolumn{3}{c}{CIRR} & \multicolumn{3}{c}{Refined-CIRR} \\
            \cmidrule(lr){3-5}\cmidrule(lr){6-8}
             \headstrut & & R@1 & R@5 & R@10 & R@1 & R@5 & R@10 \\
            \midrule
            (a) & Full Factor & 56.84 & 84.87 & 91.81 & 69.20 & 91.47 & 95.24 \\
            (b) & Mean Factor & 54.87 & 83.21 & 90.53 & 67.21 & 90.30 & 94.68 \\
            (c) & Mean Factor$^{\#}$ & 55.69 & 84.36 & 91.37 & 68.25 & 91.60 & 95.42 \\
            (d) & \eos & 56.12 & 84.58 & 91.61 & 68.84 & 90.96 & 94.93 \\
            \rowcolor{red!20}
            (e) & Efficient Rerank & 56.60 & 84.87 & 91.78 & 68.99 & 91.29 & 95.42 \\
            \bottomrule
        \end{tabular}
    \vspace{-15pt}
\end{table}
\begin{table*}[t]
    \centering
    \vspace{-5pt}
    \caption{\textbf{Ablation study.}
    Imag., NCE, and Distill represent $\mathcal{L}_{\mathrm{imag}}$~(Eq.~\ref{eq:recon}), factor-level InfoNCE, and attention distillation~(Eq.~\ref{eq:attn}), respectively.
    ERR represents efficient reranking~(Eq.~\ref{eq:inference_short}) and Post denotes the post-training stage with augmented triplets.
    R$_s$@N denotes Recall$_{\mathrm{subset}}$@N.
    }
    \label{tab:ablation}
    \vspace{-10pt}
    \setlength{\aboverulesep}{0pt}   
    \setlength{\belowrulesep}{0pt}
    \renewcommand{\arraystretch}{0.95} 
    \newcommand\headstrut{\rule[-1.1ex]{0pt}{3.5ex}}  
    \setlength{\tabcolsep}{3.15pt}
    \footnotesize
    \begin{tabular}{c|ccccc|ccccc|cc|ccccc|cc}
        \toprule
        \headstrut & \multirow{2}{*}[-0.4ex]{Imag} & \multirow{2}{*}[-0.4ex]{NCE}
        & \multirow{2}{*}[-0.4ex]{Distill} & \multirow{2}{*}[-0.4ex]{ERR} & \multirow{2}{*}[-0.4ex]{Post}
        & \multicolumn{5}{c|}{CIRR test} & \multicolumn{2}{c|}{FIQ} & \multicolumn{5}{c|}{Refined-CIRR} & \multicolumn{2}{c}{Refined-FIQ} \\
        \cmidrule(lr){7-11} \cmidrule(lr){12-13} \cmidrule(lr){14-18} \cmidrule(lr){19-20}
        \headstrut & & & & &
        & R@1 & R@5 & R@10 & R$_s$@1 & R$_s$@2
        & R@10 & R@50
        & R@1 & R@5 & R@10 & R$_s$@1 & R$_s$@2 
        & R@10 & R@50 \\
        \midrule
        (a) & - &  - & - & - & - & 51.98 & 81.90 & 90.27 & 77.69 & 91.85 & 52.98 & 73.68 & 64.64 & 89.51 & 94.53 & 84.65 & 94.68 & 66.82 & 83.55 \\
        (b) & \checkmark &  - & - & - & - & 51.45 & 82.89 & 90.36 & 77.64 & 91.76 & 53.98 & 74.26 & 63.72 & 89.56 & 94.27 & 83.73 & 94.76 & 67.98 & 84.47 \\
        (c) & \checkmark & \checkmark &  - & - & - & 52.72 & 83.28 & 90.96 & 78.68 & 92.22 & 55.25 & 75.01 & 65.17 & 89.94 & 94.65 & 84.83 & 94.81 & 68.94 & 84.97 \\
        (d) & \checkmark & - & \checkmark  & - & - & 52.53 & 82.84 & 90.82 & 78.15 & 92.02 & 54.69 & 75.36 & 64.92 & 89.74 & 94.53 & 84.14 & 94.78 & 68.40 & 84.88 \\
        (e) & \checkmark & \checkmark & \checkmark & - & - & 53.28 & 83.42 & 91.18 & 78.94 & 92.51 & 55.98 & 75.94 & 65.43 & 90.17 & 94.83 & 84.73 & 94.86 & 69.19 & 85.50 \\
        (f) & \checkmark & \checkmark & \checkmark & \checkmark & - & 54.75 & 83.86 & 91.61 & 80.84 & 93.18 & 56.67 & 76.19 & 66.55 & 90.33 & 95.11 & 85.54 & 95.47 & 69.78 & 85.67 \\
        (g) & - &  - & - & - & \checkmark & 53.69 & 83.21 & 91.01 & 80.63 & 92.99 & 54.84 & 75.12 & 66.42 & 90.19 & 95.21 & 86.33 & 95.67 & 68.76 & 84.60 \\
        (h) & \checkmark & \checkmark & \checkmark & \checkmark & \checkmark & 56.60 & 84.87 & 91.78 & 81.86 & 93.78 & 57.71 & 76.87 & 68.99 & 91.29 & 95.42 & 87.45 & 96.16 & 71.13 & 86.41 \\
        \bottomrule
    \end{tabular}
    \vspace{-5pt}
\end{table*}

\subsection{Ablation Study}
\label{sec:ablation}
In Tab.~\ref{tab:ablation}, we conduct component ablation study.
Introducing factorized latents with the target imagination alone~(b) does not consistently improve retrieval: while it improves R@10 on FashionIQ and Refined-FashionIQ by 1.00 and 1.16, and R@5 on CIRR by 0.99, it slightly lowers R@1 on CIRR.
We attribute this to the gap between generation and discrimination: target reconstruction structures the latents into semantically factorized components, which already benefits FashionIQ where modifications are confined to localized attributes, but it provides no explicit alignment metric for ranking, so the decomposition alone is insufficient for precise top-1 retrieval on open-domain CIRR.
Yet, as it lays a robust foundation for fine-grained query decomposition, we find that bridging the gap with factor-level InfoNCE~(c) and incorporating the attention distillation loss~(e) effectively enhances the retrieval performance by transferring localized decomposition into the MLLM's deeper attention layer.
Subsequently, incorporating our ERR in (f) delivers a prominent gain of 1.17 and 0.97 in Avg. on CIRR and Refined-CIRR, respectively. In contrast, applying the same reranking to latents trained without any auxiliary objective changes Avg. by only $-0.37$ and $+0.24$, indicating that the gain does not stem from merely fusing an additional vector but from the structure imposed by our objectives. 
We note that this gain comes at a marginal computational overhead, as the factorized latents are generated within the same forward pass~($+0.17$ ms per query, $<\!0.05\%$ overhead; Appendix C).
Lastly, integrating the post-training stage~(h) achieves additional gains, demonstrating the effectiveness of mitigating the noise arising from the sparse supervision of CIR datasets.
To verify that these gains are not attributed to the augmentation alone, we also apply the post-training stage directly to the baseline~(g).
While post-training alone improves the baseline, our full model~(h) further surpasses it by 2.91 and 2.57 in R@1 on CIRR and Refined-CIRR, and by 2.87 and 2.37 in R@10 on FashionIQ and Refined-FashionIQ, respectively.
Also, the gains of our components remain comparable with and without post-training~((a)$\rightarrow$(f) vs.\ (g)$\rightarrow$(h)), indicating that \ours and augmentation are complementary.
Results of post-training with ReCALL's augmented data are provided in Appendix~F.1.

\begin{figure}[t]
    \centering
    \begin{subfigure}[t]{0.95\linewidth}
        \centering
        \includegraphics[width=\linewidth]{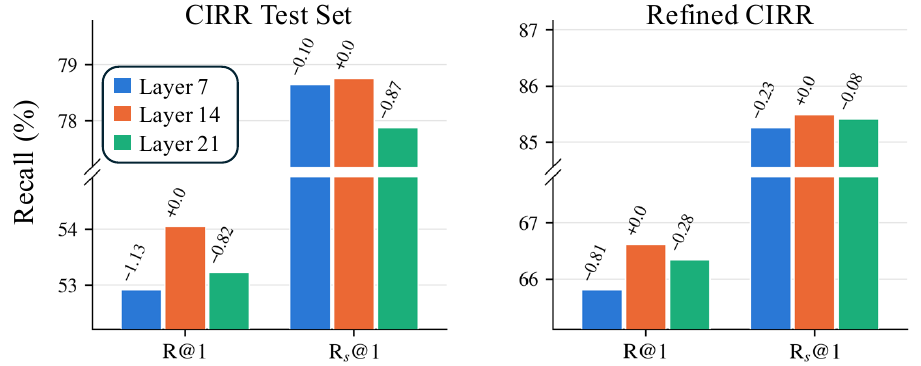}
        \vspace{-15pt}
        \caption{Impact of imagination layer.}
        \label{fig:imag_layer}
    \end{subfigure}
    \begin{subfigure}[t]{\linewidth}
        \centering
        \vspace{5pt}
        \includegraphics[width=\linewidth]{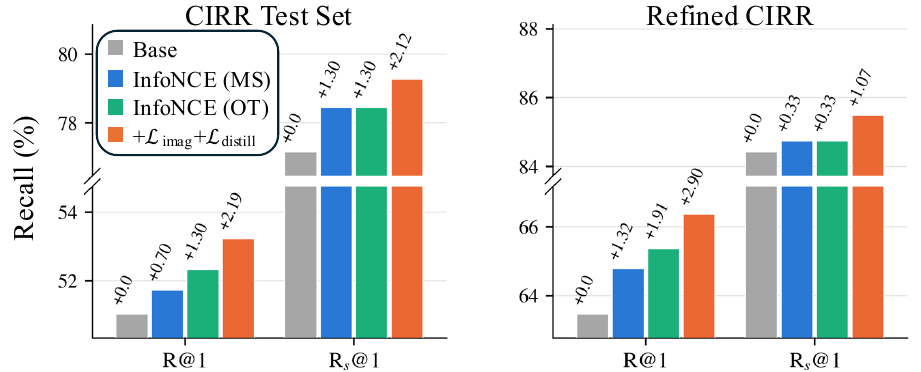}
        \vspace{-15pt}
        \caption{Impact of target imagination.}
        \label{fig:target_imag}
    \end{subfigure}
    \vspace{-5pt}
    \caption{   
    \textbf{Studies on imagination layer and comparison against NCE-only baselines with different matching metrics.}
    MS and OT denote maximum similarity and optimal transport applied to latents, respectively, without the imagination objective for factorization. 
   }
    \label{fig:furtherstudies}
    \vspace{-10pt}
\end{figure}


\begin{figure*}
    \centering
    \vspace{-11pt}
    \includegraphics[width=0.95\textwidth]{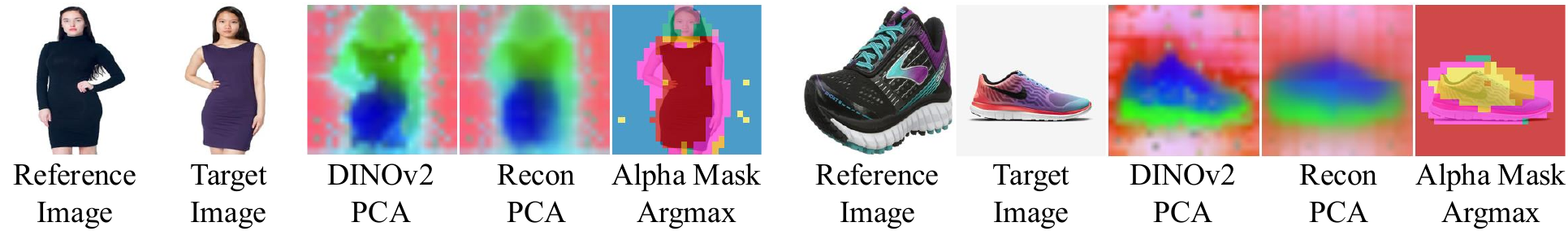}
    \vspace{-7pt}
    \caption{
    \textbf{Qualitative analysis of target imagination.}
    For each composed query, we visualize target DINOv2 features~(DINOv2 PCA) and features imagined by our decoder from the query~(Recon PCA), projected onto a joint PCA basis.
    Alpha Mask Argmax shows, for each spatial position $p$, the factorized latent with the largest decoder alpha weight, \ie, $\arg\max_k \bar{\mathcal{M}}_{k,p}$, where each color denotes a different latent.
    On top of the coarse layout from the target prior, fine details are recovered by the latents, jointly reconstructing the target structure.
    }
    \label{fig:reconpca}
    \vspace{-11pt}
\end{figure*}
\begin{figure}
    \centering
    \includegraphics[width=0.95\linewidth]{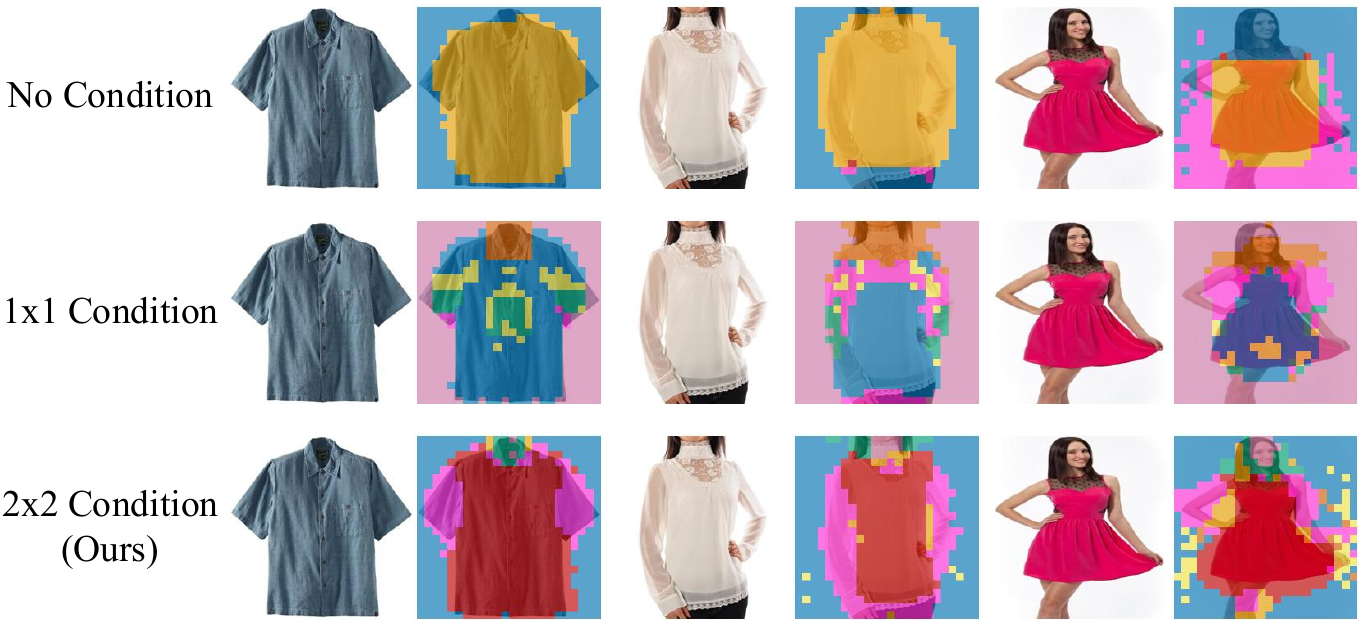}
    \vspace{-8pt}
    \caption{
        \textbf{Impact of layout prior in target imagination.}
        Three rows compare factor decomposition across different layout prior configurations: (top) without a layout prior, (middle) with a $1\times1$ layout prior, and (bottom) with our default $2\times2$ layout prior. 
        While the unconditioned model collapses into coarse global semantics, introducing the $2\times2$ layout prior guides the latents to isolate semantically editable parts.
    }
    \label{fig:condition_effect}
    \vspace{-14pt}
\end{figure}

\subsection{Analysis and Discussion}
\label{sec:furtherstudy}
\textbf{Imagination Layer Selection.}
Determining the MLLM layer $\ell$ for target imagination governs a trade-off; when the decoder is attached too early, the factorized latents may lack sufficient contextual fusion of multimodal inputs to imagine the target, while attaching it too late leaves insufficient subsequent layers to transform the decomposed factors into a discriminative retrieval embedding.
In Fig.~\ref{fig:imag_layer}, we evaluate decoder placements at layers $\ell \in \{7, 14, 21\}$ of the 28-layer MLLM, and find that $\ell=14$ yields the best result, which aligns with recent findings~\citep{jiang2025devils, yoon2025viral}.

\noindent\textbf{Impact of Target Imagination.}
To isolate the effect of target imagination, we compare against variants that train the latents solely with InfoNCE, matched by either maximum similarity~(MS)~\citep{xiao2026metaembed} or OT, in Fig.~\ref{fig:target_imag}.
While these variants outperform the na\"ive baseline, we observe that target imagination brings a further gain by enforcing the latents to decompose and structure the target scene rather than relying on the contrastive signal alone.
Also, OT matching consistently outperforms MS matching strategy on both benchmarks, supporting our choice of OT for factor-level matching.

\noindent\textbf{What Does the Decoder Imagine?}
Fig.~\ref{fig:reconpca} compares the imagined features $\hat{F}^{q}$ against the GT target features $F$, using joint PCA basis.
On the query side, the imagined features reconstruct the overall structure of the target: while the layout prior supplies only the coarse layout, the factorized latents recover fine-grained details such as object silhouettes.
The rightmost column visualizes $\arg\max_{k}\bar{\mathcal{M}}_{k,p}$, demonstrating that each factorized latent contributes to reconstructing semantically coherent regions.
This suggests that each latent specializes in a coherent semantic factor of the target.
Additional visualizations are in Appendix A.

\noindent\textbf{Impact of Target Layout Prior.}
As described in Sec.~\ref{sec:method:attributelatents}, the decoder is conditioned on a coarse target layout prior.
Fig.~\ref{fig:condition_effect} visualizes the decomposition under different prior resolutions.
Without the layout prior, the latents must also encode the global layout, which dominates the reconstruction objective and causes them to collapse into a few coarse regions.
In contrast, when the prior is injected, we found that the factorized latents are encouraged to focus on fine-grained visual structures, decomposing the target into localized parts such as arms, legs, and necklines.
We set $h=2$ so that the layout prior provides only the coarse layout while leaving the localized factors to the latents.

\noindent\textbf{Generalization across Backbones.}
We further apply \ours to stronger MLLM backbones in Tab.~\ref{tab:backbone}. 
Specifically, we observe that upgrading the backbone from Qwen2.5-VL-7B to Qwen3-VL-8B improves R@1 by 2.65 points on the CIRR test set, exceeding the 1.60 points gain reported by ReCALL~\citep{RECALL} for the same backbone transition. 
This demonstrates that \ours scales more effectively with improved backbone capabilities.
The performance gains further extend to Qwen3-VL-Embedding-8B, a backbone already specialized for retrieval, where \ours achieves top performance across most metrics.


\noindent\textbf{Do Different Edits Engage Different Latents?}
To isolate the effect of the modification text, we encode reference images with and without a caption and measure how much each latent changes.
We use FashionIQ, where each query has two modification texts for the same reference--target pair.
Since the two texts often describe different aspects of the change, we use Qwen3.5-9B~\citep{qwen35blog} to select the pairs that describe the same edit, which yields 833 pairs.
After subtracting each latent's average change, the two captions of these pairs agree strongly on which latents change the most~(correlation $0.67$), whereas a caption from another query does not~($-0.03$).
This indicates that which latents respond depends on the content of the edit rather than on the mere presence of text.

\section{Conclusion}
We present \ours, which exploits the MLLM's compositional understanding capability by decoupling compositional query comprehension from discriminative embedding learning.
To structure query comprehension, we insert factorized latents before \eos and train them to imagine the target beyond a coarse layout prior without dense annotations.
We then bridge generative imagination and discriminative retrieval with a factor-level InfoNCE loss and attention distillation.
At inference, \eos provides a global embedding for initial retrieval, while the mean-pooled latents, produced in the same forward pass, rerank a short candidate list at negligible cost.
Experiments show that \ours outperforms prior work and further improves with stronger MLLM backbones, suggesting that factor decomposition provides an effective foundation for forming retrieval embeddings in MLLM-based CIR.
\begin{table}[t]
    \centering
    \vspace{0pt}
    \caption{\textbf{\ours with different MLLM backbones.}
    ``Avg.'' follows Tab.~\ref{tab:cirr} and Tab.~\ref{tab:refined_results_combined}, \ie, (R@5 + R$_s$@1)/2 on CIRR test and (R@1 + R$_s$@1)/2 on Refined-CIRR.
    }
    \label{tab:backbone}
    \vspace{-5pt}
    \setlength{\aboverulesep}{0pt}
    \setlength{\belowrulesep}{0pt}
    \renewcommand{\arraystretch}{1.0}
    \newcommand\headstrut{\rule[-1.1ex]{0pt}{3.5ex}}
    \newcommand\rotlabel[1]{\rotatebox[origin=c]{90}{\shortstack{#1}}}
    \setlength{\tabcolsep}{1.4pt}
    \scriptsize
    \begin{tabular}{c|l|cccccccc}
        \toprule
        \headstrut & Backbone
        & R@1 & R@5 & R@10 & R@50 & R$_s$@1 & R$_s$@2 & R$_s$@3 & Avg. \\
        \hline
        \multirow{3}{*}{\rotlabel{CIRR\\test}}
        & Qwen2.5-VL-7B & 56.60 & 84.87 & 91.78 & 98.41 & 81.86 & 93.78 & 97.61 & 83.37 \\
        & Qwen3-VL-8B & 59.25 & 86.10 & 91.93 & 98.36 & 83.54 & 94.53 & 97.74 & 84.82 \\
        & Qwen3-VL-Emb-8B & \textbf{60.43} & \textbf{86.99} & \textbf{93.35} & \textbf{98.70} & \textbf{84.15} & \textbf{94.82} & \textbf{97.81} & \textbf{85.57} \\
        \hline
        \multirow{3}{*}{\rotlabel{Refined\\CIRR}}
        & Qwen2.5-VL-7B & 68.99 & 91.29 & 95.42 & 99.06 & 87.45 & 96.16 & 98.63 & 78.22 \\
        & Qwen3-VL-8B & 71.31 & 92.29 & 95.72 & \textbf{99.16} & \textbf{89.33} & 96.59 & \textbf{98.88} & 80.32 \\
        & Qwen3-VL-Emb-8B & \textbf{73.60} & \textbf{93.02} & \textbf{96.41} & 99.13 & 88.95 & \textbf{97.12} & \textbf{98.88} & \textbf{81.28} \\
        \bottomrule
    \end{tabular}
    \vspace{-18pt}
\end{table}

\textbf{Limitations.}
Despite its effectiveness, \ours has several limitations.
First, using a fixed $K$ ignores the input complexity since no explicit supervision specifies the number of semantic factors.
Exploring query-adaptive latents and matching schemes across varying set sizes remains for future work.
Second, while full-factor retrieval achieves higher accuracy, we mean-pool the latents for efficient reranking, unexploiting part of the fine-grained information.
Incorporating Matryoshka representation learning~\citep{kusupati2022matryoshka} into factor representations may preserve accuracy of full-factor matching near single-vector storage costs.
Lastly, our evaluation is limited to CIR datasets; thus, generalizability as a unified multimodal retriever requires further verification.
\newpage



\renewcommand{\thetable}{A\arabic{table}}   
\renewcommand{\thefigure}{A\arabic{figure}}
\renewcommand{\theequation}{A\arabic{equation}}
\setcounter{section}{0}
\setcounter{table}{0}
\setcounter{figure}{0}
\setcounter{equation}{0}


\appendix
\begin{figure*}[t]
    \centering
    \includegraphics[width=1.\textwidth]{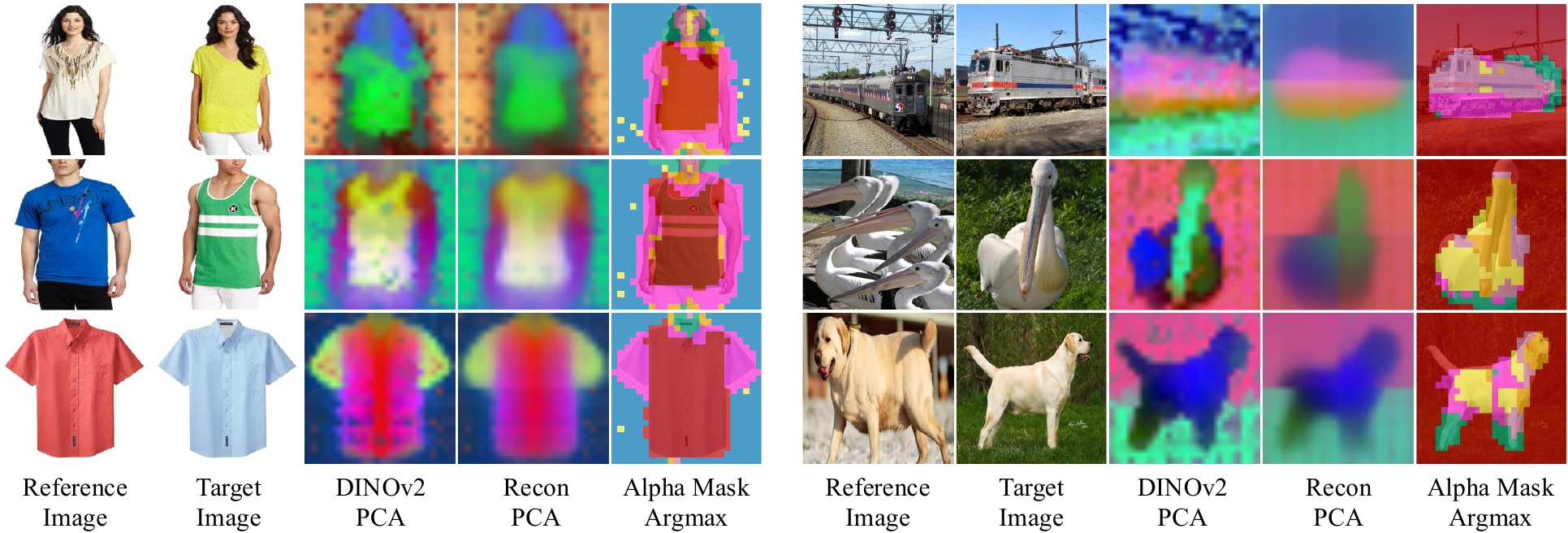}
    \caption{
        \textbf{Additional Qualitative Analysis of Target Imagination.}
    }
    \label{fig:reconpcaqualsupp}
\end{figure*}
\section{Additional Visualizations for Imagination}
\label{sec.app.addiimag}
In this section, we present additional results for target imagination in Fig.~\ref{fig:reconpcaqualsupp}. 
Following the protocol in Sec. 4.4, we jointly apply PCA to the ground-truth DINOv2~\citep{oquab2023dinov2} feature maps and our imagined representations for visualization. 
Although the shared color mapping between both visualized images is partly facilitated by the target layout prior, we remark that this prior provides only a highly coarse, average-pooled $2\times2$ spatial layout. 
Consequently, the reconstruction of details~(\textit{e.g.,} geometry and position) suggests that our factorized latents capture rich spatial positioning and object geometry, allowing the decoder to synthesize the spatial maps across local regions.

\begin{table*}[t]
\centering
\caption{\textbf{Diversity analysis on factorized latents.} 
``Within instance'' denotes the mean cosine similarity across all latent pairs within the same image, whereas ``Across instances'' denotes the mean cosine similarity between latents of different images.
Further, we report prototype-removed diversity which is computed by subtracting the dataset-wide per-latent mean vector~(computed over the CIRR dataset) from each latent to isolate diversity from MLLM residual biases.
}
\label{tab:slot_diversity}
\small
\begin{tabular}{c|l|cc|c}
\toprule
& \multirow{2}{*}{Methods} & \multicolumn{2}{c|}{Raw Similarity} & Prototype-removed \\
 & & within instance & across instances & within instance \\
\midrule
a & Learnable Latents only (no auxiliary loss)            & 0.606 & 0.408 & 0.599 \\
b & Latents with InfoNCE                           & 0.764 & 0.548 & 0.841 \\
c & Ours & 0.586 & 0.426 & 0.548 \\
\bottomrule
\end{tabular}
\end{table*}

\section{Diversity of Factorized Latents}
To examine whether the factorized latents remain distinct from one another, we measure the cosine similarity between factorized latents both within the same instance and across different instances~(Tab.~\ref{tab:slot_diversity}). 
Specifically, we compare our approach against two baseline variants: one where learnable latents are prepended to the \eos token without additional constraints, and another where the factor-level InfoNCE objective is introduced.

As discussed in Sec. 3.3, applying the factor-level InfoNCE alone without the imagination objective $\mathcal{L}_{\mathrm{imag}}$~(Tab.~\ref{tab:slot_diversity}~(b)) raises the within-instance similarity from 0.606 to 0.764, indicating that the latents collapse toward a single instance-level representation.
This is because the retrieval objective only requires the latents to be collectively discriminative across instances, providing no incentive for individual latents to take on distinct roles.
Furthermore, this representation collapse becomes even more pronounced when the per-latent mean vector~(computed across the dataset) is subtracted from each latent, indicating that the latents share instance-specific rather than factor-specific information.
In contrast, when the same InfoNCE is combined with our imagination and distillation objectives~(c), the within-instance similarity stays at 0.586~(0.548 after mean removal), showing that these objectives counteract the collapse induced by the retrieval loss.

Note that the latents trained without any auxiliary objective~(a) show a similarity comparable to ours~(0.606 and 0.599). 
Low similarity alone, however, does not imply that the latents are useful: these latents receive no direct supervision, and they do not benefit retrieval.
When the latents in (a) are used for efficient reranking, they achieve an average of 73.45 on Refined-CIRR and 79.60 on the CIRR test set~(Tab.~\ref{tab:rerank_control}~(c)), falling below the baseline without latents~(74.65 and 79.80), where the averages follow Tab.~1 and Tab.~2, \ie, (R@1 + R$_{\text{subset}}$@1)/2 and (R@5 + R$_{\text{subset}}$@1)/2, respectively.

\begin{table*}[t]
\centering
\small
\renewcommand{\arraystretch}{0.95}  
\setlength{\tabcolsep}{3.6pt} 
\caption{\textbf{Per-query inference cost on CIRR validation set~(2,297 gallery images). }
    We report the number of vectors indexed per gallery image, the scoring cost~(similarity matching), and the total inference latency for both single and batched queries.
    Total inference latency involves the scoring cost and the encoding cost of the query-side, while the cost of encoding offline gallery~(target-side) is excluded.
    $\Delta$ indicates the latency overhead relative to the \eos-only baseline.
    }
\label{tab:efficiency}
\begin{tabular}{l c c c c c c c}
\toprule
\multirow{3}{*}{Retrieval Unit} & \multirow{3}{*}{\begin{tabular}[c]{@{}c@{}}Num\\Vectors\end{tabular}} & \multicolumn{2}{c}{Scoring} & \multicolumn{4}{c}{Total} \\
\cmidrule(lr){3-4} \cmidrule(lr){5-8}
 & & Single (bs=1) & Batched (bs=32) & \multicolumn{2}{c}{Single (bs=1)} & \multicolumn{2}{c}{Batched (bs=32)} \\
\cmidrule(lr){3-3} \cmidrule(lr){4-4} \cmidrule(lr){5-6} \cmidrule(lr){7-8}
 & & ms/q & ms/q & ms/q & $\Delta$ & ms/q & $\Delta$ \\
\midrule
  \eos only & 1  & 0.06 & 0.004 & 353.1 & --      & 163.8 & --       \\
  + Efficient Rerank           & 2  & 0.23 & 0.007 & 353.3 & +0.17ms & 163.8 & +0.003ms \\
  Full Factor  & 8 & 3.99 & 0.31  & 357.1 & +3.93ms & 164.1 & +0.31ms  \\
\bottomrule
\end{tabular}
\end{table*}

\section{Inference Cost}
\label{sec:efficiency}
A natural concern with multi-vector representations is that they trade retrieval quality for memory footprint and query latency~\citep{khattab2020colbert, xiao2026metaembed}.
To address this, Tab.~\ref{tab:efficiency} quantifies the inference overhead on the CIRR validation set~(2,297 gallery images) using a single NVIDIA RTX PRO 5000 Blackwell GPU. 
Crucially, the forward pass of the query encoder dominates the end-to-end latency pipeline, requiring 353ms per query at batch size 1 and an average of 164ms per query at batch size 32. 
In contrast, scoring the gallery using only the \eos embedding requires only 0.06ms per query. 
Our efficient reranking procedure, which re-scores the top-100 shortlist with the sum of the cosine similarities from each \eos and mean factorized vector, adds one extra vector per image to the index and 0.17ms per query at batch size 1~(0.003ms amortized), which only induces less than 0.05\% overhead on the total latency. 
We adopt this design as our default inference configuration, as its storage footprint remains comparable to, or slightly smaller than that of recent CIR methods that maintain multi-vector galleries~\citep{ConeSep, TME, CCIN, QuRe, CaLa, SPRC}.
When storage and compute budgets permit, however, the full latent variant can be preferred, as full OT matching offers slightly higher R@1~(Tab.~4).
Note that its time overhead remains marginal at the scale of the CIRR validation gallery, but both its storage~($K$ vectors per image) and its scoring cost grow linearly with the gallery size, whereas efficient reranking only rescores a fixed top-100 shortlist.

\section{Additional Implementation Details}
In this section, we elaborate additional implementation details.
While we use base learning rates of $2\times 10^{-5}$ and $4\times 10^{-5}$ to train DoRA modules attached to the pretrained backbone parameters for CIRR and Fashion-IQ, respectively, we apply $4\times 10^{-4}$ as the learning rate to modules such as the factorized latent and imagination decoder since our newly introduced components are trained from scratch.
For the training schedule, we use 3,000, 500, and 350 steps across the three stages for CIRR, and 3,000, 350, and 350 steps for FashionIQ. 
While Stage~1 could technically be trained jointly with the later stages, the competing dynamics between $\mathcal{L}_{\mathrm{imag}}$ and $\mathcal{L}_{\mathrm{NCE}}$ slow down the formation of factorized latents, requiring substantially longer training to converge.
Isolating Stage~1 to train only the decoder and the latents with $\mathcal{L}_{\mathrm{imag}}$ removes this competition and further allows bypassing the upper layers of the MLLM backbone, which reduces the cost of this stage: Stage 1 completes in approximately 13 hours on 4× RTX PRO 5000 Blackwell GPUs.
Then, subsequent training stages were conducted using 2$\times$ RTX PRO 6000 Blackwell GPUs using PyTorch FSDP.
Images are resized within the processor to between $3{,}136$ and $147{,}456$ pixels, with input sequences truncated at $512$ tokens. 
\section{Optimal Transport for Matching Score between Factorized Sets}
\label{sec.ot}
To measure the similarity between two unordered factorized sets, we used Optimal Transport~(OT) to find the best possible alignment between them. 
Given the query and target factorized latents $\mathbf{U}^q = [u^q_1, \dots, u^q_K]$ and $\mathbf{U}^t = [u^t_1, \dots, u^t_K]$, we first compute a pairwise cosine similarity matrix $\mathbf{R} \in \mathbb{R}^{K \times K}$, where $\mathbf{R}_{k,m} = \cos(u^q_k, u^t_m)$ is the cosine similarity between the $k$-th query factor and the $m$-th target factor.
Based on this similarity matrix $\mathbf{R}$, the OT matching score is defined as:
\begin{equation}
\begin{aligned}
  &s_{\mathrm{OT}}(\mathbf{U}^q, \mathbf{U}^t) = \big\langle \mathbf{P}^{\star}, \mathbf{R} \big\rangle, \\
  &\mathbf{P}^{\star} = \arg\max_{\mathbf{P} \in \Pi(\frac{1}{K}\mathbf{1}, \frac{1}{K}\mathbf{1})}
  \big\langle \mathbf{P}, \mathbf{R} \big\rangle + \varepsilon \mathcal{H}(\mathbf{P}),
\label{eq:ot}
\end{aligned}
\end{equation}
where $\mathbf{P}^{\star}$ is the optimal transport plan that maximizes the overall alignment score and $\langle \cdot, \cdot \rangle$ denotes the Frobenius inner product.
$\Pi(\tfrac{1}{K}\mathbf{1}, \tfrac{1}{K}\mathbf{1})$ denotes the set of transport plans whose rows and columns each sum to $1/K$, so that every query and target factor contributes equally.
The entropy regularization $\mathcal{H}(\mathbf{P}) = -\sum_{k,m} \mathbf{P}_{k,m} \log \mathbf{P}_{k,m}$ allows $\mathbf{P}^{\star}$ to be computed efficiently with a few Sinkhorn iterations, serving as a differentiable surrogate for one-to-one matching.
We set the $\varepsilon = 0.1$ and run 30 Sinkhorn iterations.
Since the elements of $\mathbf{P}^{\star}$ sum to 1, $\langle \mathbf{P}^{\star}, \mathbf{R} \rangle$ represents the average matching score in $[-1, 1]$, which is subsequently used as the factor-level similarity $\mathbf{S}^{f}$ in the first term of Eq.~5.
This encourages the factorized set of a query to align with that of its target while repulsing the sets of non-matching targets.

\newcommand{\Ir}{I_{r}}
\newcommand{\It}{I_{t}}
\newcommand{\Ih}{I_{h}}
\newcommand{\Tm}{T}
\newcommand{\Tt}{\Acute{T}}

\section{Post-training with MLLM Augmentation}
\label{sec:aug}
A typical challenge in CIR is that the modification text often underdetermines the target image. 
In other words, instructions are rarely exhaustive; they specify what should change while leaving everything else implicit. 
Consequently, a single composed query can often be satisfied by multiple candidate images in the gallery. 
To address this, prior works have leveraged LLM/MLLMs to sharpen the query itself to disambiguate the instruction~\citep{CCIN, garderes2026fire} or enlarged the clean paired triplets~\citep{zhang2024magiclens, zhou2025megapairs, huynh2025collm, RECALL}. 
Similarly, we follow ReCALL~\citep{RECALL} to augment training triplets by pairing each reference image with its mined hard negative instances, and then finetune the model on the resulting data~(comprising the reference image, its hard negative, and a generated modification text). 
Operating on \ours after Stage 2, our augmentation pipeline consists of two phases: hard negative mining and contrastive captioning. 
During the mining phase, we follow ReCALL to extract hard negative instances $\Ih$ for each reference image $\Ir$ that ranks above the ground truth $\It$, which are then passed to the contrastive captioning phase.

\subsection{Contrastive Captioning}
\label{sec:aug:rewriting}
Following ReCALL~\citep{RECALL}, we convert each mined hard negative $\Ih$ into a new positive triplet $(\Ir, \Tt, \Ih)$, where an MLLM writes a minimal modification text $\Tt$ that transforms $\Ir$ into $\Ih$.
In practice, however, we found this pipeline to be both noisy and costly.
Since mined hard negatives frequently serve as valid targets for $(\Ir, \Tm)$ themselves and the captioner observes only $\Ir$ and $\Ih$, we observed that the captioner has no cue to separate $\Ih$ from $\It$.
As a result, it often returned $\Tm$ unchanged, producing duplicate queries with conflicting positives~(9\% and 11\% of the samples that survived ReCALL's VQA filter on CIRR and FashionIQ).
This pipeline was also inefficient: the local MLLM captioner frequently violated the required output format, with 37\%~(CIRR) and 56\%~(FashionIQ) of instances failing on the first pass and 9\% and 12\% remaining unresolved even after six retry rounds, and the subsequent VQA filter added another full inference pass.

\begin{figure}[t]
    \centering
    \vspace{-0.0cm}
    \begin{subfigure}[t]{\linewidth}
        \centering
        \includegraphics[width=\linewidth]{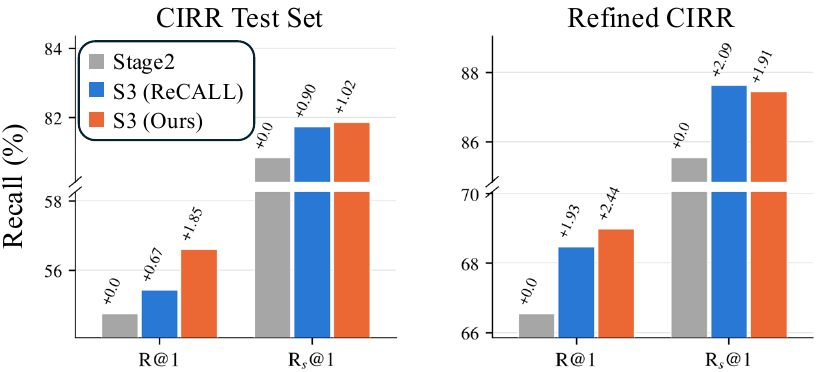}
        \vspace{-5pt}
        \caption{Impact of Stage3 Post-training on CIRR dataset.}
        \label{fig:post_cirr}
    \end{subfigure}
    \begin{subfigure}[t]{\linewidth}
        \centering
        \includegraphics[width=\linewidth]{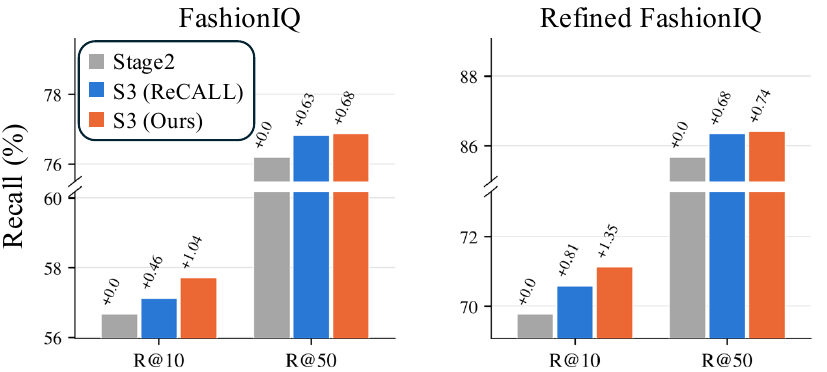}
        \vspace{-5pt}
        \caption{Impact of Stage3 Post-training on FashionIQ Dataset.}
        \label{fig:post_fiq}
    \end{subfigure}
    \caption{   
    \textbf{Impact of augmentation strategies for post-training.}
        Starting from our model before post-training~(Stage 2), we compare post-training with triplets augmented by the two-image prompting of ReCALL~\citep{RECALL} and by our contrastive captioning.
        Numbers above the bars denote the absolute gains over stage 2 baseline.
    }
    \label{fig:augstudies}
\end{figure}

\begin{table*}[t]
    \centering
    \caption{\textbf{Results on additional benchmarks.}
    enhFashionIQ and Shoes report Recall@N with FashionIQ-trained models;
    CIRCO reports mAP@N on the test server with CIRR-trained models.
    Best results are in bold.}
    \label{tab:more_benchmarks}
    \setlength{\aboverulesep}{0pt}
    \setlength{\belowrulesep}{0pt}
    \renewcommand{\arraystretch}{1.15}
    \newcommand\headstrut{\rule[-1.1ex]{0pt}{3.5ex}}
    \setlength{\tabcolsep}{5pt}
    \footnotesize
    \begin{tabular}{l|ccc|ccc|cccc}
        \toprule
        \headstrut \multirow{2}{*}{Method}
        & \multicolumn{3}{c|}{enhFashionIQ}
        & \multicolumn{3}{c|}{Shoes}
        & \multicolumn{4}{c}{CIRCO (mAP)} \\
        \cmidrule(lr){2-4} \cmidrule(lr){5-7} \cmidrule(lr){8-11}
        \headstrut
        & R@1 & R@10 & R@50
        & R@1 & R@10 & R@50
        & @5 & @10 & @25 & @50 \\
        \midrule
        Baseline
            & 38.17 & 75.59 & 92.03
            & 16.81 & 44.07 & 70.70
            & 28.52 & 28.98 & 31.26 & 32.26 \\
        ReCALL
            & 41.97 & 77.94 & 92.65
            & 18.06 & 47.98 & 73.54
            & 28.43 & 29.01 & 31.25 & 32.26 \\
        \midrule
        \ours (\eos)
            & 44.72 & 80.95 & 93.90
            & 18.68 & 48.04 & \textbf{74.11}
            & \textbf{29.16} & \textbf{29.41} & \textbf{31.67} & \textbf{32.73} \\
        \ours (+ ERR)
            & \textbf{45.12} & \textbf{81.10} & \textbf{93.95}
            & \textbf{19.19} & \textbf{48.32} & 73.82
            & 28.57 & 28.92 & 31.18 & 32.20 \\
        \bottomrule
    \end{tabular}
\end{table*}

We thus streamline the pipeline with two modifications, while keeping the remaining instructions of ReCALL's captioning prompt unchanged.
First, we employ contrastive captioning strategy, which additionally presents the original target $\It$ alongside $\Ir$ and $\Ih$, and instruct the model to describe $\Ih$ such that the text no longer applies to $\It$.
Note that $\It$ serves purely as a conditioning signal and is omitted from the augmented triplet.
Second, we replace the local captioner with GPT-5.6-luna~\citep{gpt56luna} via the batch API, which can follow both the contrast instruction and the output format in a single pass, removing the retry loop and separate verification stage.

Compared with the ReCALL's two-image prompting strategy on 100 sampled mined queries per dataset, we found that our contrastive captioning successfully yields an edited modification text while the two-image prompting strategy produced 27.0\%~(CIRR) and 33.0\%~(FashionIQ) of modification texts unchanged.
Furthermore, among the edited queries, we found that 30.1\%~(CIRR) and 10.5\%~(FashionIQ) of the modification texts  rewritten by the two-image prompting strategy were judged by GPT-5.6-luna to describe the original target $\It$ better than the mined target $\Ih$, while our contrastive captioning reduced this reversal rate to 7.0\% and 0.0\%, respectively.
Since this check relies on the same model as the captioner and on a small sample, we regard it as a sanity check rather than a rigorous quality evaluation.
For downstream retrieval task, Fig.~\ref{fig:augstudies} shows that post-training \ours with our triplets is at least on par with using ReCALL's triplets: the gains are larger on R@1~(CIRR) and R@10~(FashionIQ) and similar on the remaining metrics, with a slightly lower R$_s$@1 on Refined-CIRR.
This suggests that our streamlined pipeline, with only a single generation pass, retains the benefit of augmentation without a caption retry loop or a separate verification stage.
Other than these modifications, we followed ReCALL~\citep{RECALL} to mine five hard negatives per composed query, resulting in 57,476~(CIRR) and 56,499~(FashionIQ) captioning requests.
This incurs a one-time API cost of only \$22.82~(CIRR) and \$26.17~(FashionIQ).

\subsection{Hard Negative Filtering}
\label{sec:aug:filtering}
Hard negative mining occasionally yields near-duplicate instances that are virtually identical to the annotated target despite holding distinct identifiers in the same dataset. 
Rewriting such a pair forces the model to generate text that separates two near-identical images, teaching the retriever to incorrectly push them apart. 
To prevent this, we filter out these pairs prior to contrastive captioning based on visual feature similarity via DINOv2~\citep{oquab2023dinov2} before we query the MLLM to write the captions. 
Specifically, we encoded each image using a frozen DINOv2 ViT-B/14 backbone and mean-pooled the patch tokens into a single descriptor, discarding hard negative candidates~($\Ih$) that exhibit a cosine similarity exceeding $0.99$ with the original target~($\It$). 
This filtering step removes $743$ CIRR instances and $277$ FashionIQ instances out of 58$,$219 and 56$,$776 mined pairs, respectively.
For Stage 3, we additionally filter mislabeled original triplets: when mining with the Stage 2 model, we also record the rank of the ground-truth target, and drop the 3\% of original triplets whose ground truth is ranked lowest.

\subsection{Post-training}
Finally, for the post-training stage, we optimize our model using the total objective, described in Eq.~8, similarly to stage 2.
Note that we construct each batch for CIRR by grouping an original triplet with its MLLM-augmented counterparts following~\citep{RECALL}.
This strategy supplies high-quality in-batch hard negatives, explicitly encouraging the model to discern subtle visual differences.

\subsection{Limitation of Contrastive Captioning}
While our contrastive captioning mitigates noisy supervision in both the original and augmented triplets, a limitation remains when scaling up the augmentation.
Contrastive captioning conditions each rewrite only on the original target $\It$, so it reduces ambiguity between  $\It$ and a hard negative but not among the multiple hard negatives mined for the same composed query. 
As more hard negatives are mined per query, \textit{i.e.,} each pair of reference image $\Ir$ and modification text $\Tm$, we expect these sibling hard negatives to increasingly overlap in semantics, so that a modification text written for one may equally describe another, introducing false negatives within the augmented set.
An effective augmentation strategy for CIR therefore requires conditioning across both the original and augmented sets, so that each generated instruction captures features that distinguish its target from all other candidates of the same query.
Nevertheless, since the augmented triplets consistently improve CIR performance in our experiments, we will publicly release them to facilitate future research.

\begin{table*}[t]
    \centering
    \caption{\textbf{\ours with stronger MLLM backbones.}
    ``Baseline'' denotes the same backbone fine-tuned without \ours.
    ``Avg.'' on Refined-CIRR is (R@1 + R$_s$@1)/2, and on Refined-FashionIQ
    it is the mean over the three categories.
    Best results per backbone are in bold.}
    \label{tab:backbone2}
    \setlength{\aboverulesep}{0pt}
    \setlength{\belowrulesep}{0pt}
    \renewcommand{\arraystretch}{1.15}
    \setlength{\tabcolsep}{4pt}
    \footnotesize
    \begin{tabular}{l cccc ccc c cc cc cc cc}
        \toprule
        \multirow{3}{*}{Method}
        & \multicolumn{8}{c}{Refined-CIRR}
        & \multicolumn{8}{c}{Refined-FashionIQ} \\
        \cmidrule(lr){2-9} \cmidrule(lr){10-17}
        & \multicolumn{4}{c}{Recall@N}
        & \multicolumn{3}{c}{Recall$_{\mathrm{subset}}$@N}
        & \multirow{2}{*}{Avg.}
        & \multicolumn{2}{c}{Dress}
        & \multicolumn{2}{c}{Shirt}
        & \multicolumn{2}{c}{Top\&Tee}
        & \multicolumn{2}{c}{Avg.} \\
        \cmidrule(lr){2-5} \cmidrule(lr){6-8} \cmidrule(lr){10-11} \cmidrule(lr){12-13} \cmidrule(lr){14-15} \cmidrule(lr){16-17}
        & R@1 & R@5 & R@10 & R@50
        & R@1 & R@2 & R@3 &
        & R@10 & R@50
        & R@10 & R@50
        & R@10 & R@50
        & R@10 & R@50 \\
        \midrule
        \multicolumn{17}{l}{\textit{Qwen3-VL-8B}} \\
        Baseline
            & 68.23 & 91.45 & 95.47 & 99.13 & 86.66 & 95.72 & 98.45 & 77.45
            & 63.04 & 81.35 & 70.22 & 85.92 & 71.01 & 87.57 & 68.09 & 84.95 \\
        \ours
            & \textbf{71.31} & \textbf{92.29} & \textbf{95.72} & \textbf{99.16} & \textbf{89.33} & \textbf{96.59} & \textbf{98.88} & \textbf{80.32}
            & \textbf{67.90} & \textbf{84.70} & \textbf{73.58} & \textbf{87.92} & \textbf{75.41} & \textbf{89.63} & \textbf{72.30} & \textbf{87.42} \\
        \midrule
        \multicolumn{17}{l}{\textit{Qwen3-VL-Embedding-8B}} \\
        Baseline
            & 71.92 & \textbf{93.33} & \textbf{96.56} & \textbf{99.31} & 88.29 & 96.61 & 98.73 & 80.11
            & 67.50 & 83.70 & 73.27 & 87.08 & 75.46 & 89.74 & 72.08 & 86.84 \\
        \ours
            & \textbf{73.60} & 93.02 & 96.41 & 99.13 & \textbf{88.95} & \textbf{97.12} & \textbf{98.88} & \textbf{81.28}
            & \textbf{69.63} & \textbf{85.43} & \textbf{75.00} & \textbf{88.45} & \textbf{77.31} & \textbf{90.01} & \textbf{73.98} & \textbf{87.96} \\
        \bottomrule
    \end{tabular}
\end{table*}
\begin{table*}[t]
    \centering
    \caption{\textbf{Reranking with and without our auxiliary objectives.}
    ``Latents~(no aux.)'' inserts $K$ latents trained only through the instance-level InfoNCE on \eos, while ``Latents + aux.'' is our Stage-2 model. 
    ERR denotes efficient reranking. 
    }
    \label{tab:rerank_control}
    \setlength{\aboverulesep}{0pt}
    \setlength{\belowrulesep}{0pt}
    \renewcommand{\arraystretch}{1.15}
    \newcommand\headstrut{\rule[-1.1ex]{0pt}{3.5ex}}
    \setlength{\tabcolsep}{4.7pt}
    \footnotesize
    \begin{tabular}{c|l|ccccc|cc|ccccc|cc}
        \toprule
        \headstrut & \multirow{2}{*}{Method}
        & \multicolumn{5}{c|}{CIRR test} & \multicolumn{2}{c|}{FIQ}
        & \multicolumn{5}{c|}{Refined-CIRR} & \multicolumn{2}{c}{Refined-FIQ} \\
        \cmidrule(lr){3-7} \cmidrule(lr){8-9} \cmidrule(lr){10-14} \cmidrule(lr){15-16}
        \headstrut & & R@1 & R@5 & R@10 & R$_s$@1 & R$_s$@2
        & R@10 & R@50
        & R@1 & R@5 & R@10 & R$_s$@1 & R$_s$@2
        & R@10 & R@50 \\
        \midrule
        (a) & Baseline
            & 51.98 & 81.90 & 90.27 & 77.69 & 91.85 & 52.98 & 73.68
            & 64.64 & 89.51 & 94.53 & 84.65 & 94.68 & 66.82 & 83.55 \\
        \midrule
        (b) & Latents (no aux.)
            & 51.45 & 82.41 & 89.98 & 77.52 & 91.25 & 54.29 & 74.23
            & 62.78 & 89.38 & 94.70 & 83.63 & 94.88 & 67.84 & 84.34 \\
        (c) & (b) + ERR
            & 51.54 & 81.90 & 89.88 & 77.30 & 91.64 & 52.30 & 73.27
            & 63.06 & 88.98 & 94.58 & 83.83 & 94.53 & 67.28 & 84.29 \\
        \midrule
        (d) & Latents + aux.
            & 53.28 & 83.42 & 91.18 & 78.94 & 92.51 & 55.98 & 75.94
            & 65.43 & 90.17 & 94.83 & 84.73 & 94.86 & 69.19 & 85.50 \\
        (e) & (d) + ERR
            & \textbf{54.75} & \textbf{83.86} & \textbf{91.61} & \textbf{80.84} & \textbf{93.18}
            & \textbf{56.67} & \textbf{76.19}
            & \textbf{66.55} & \textbf{90.33} & \textbf{95.11} & \textbf{85.54} & \textbf{95.47}
            & \textbf{69.78} & \textbf{85.67} \\
        \bottomrule
    \end{tabular}
\end{table*}
\section{Evaluation on Additional Benchmarks}
\label{sec:supp_benchmarks}
While the evaluation in the main paper focuses on CIRR and FashionIQ, we provide a preliminary study of how \ours transfers to other CIR benchmarks. 
Specifically, we further evaluate on enhFashionIQ~\cite{enhfiq}, Shoes~\cite{shoes1, shoes2}, and CIRCO~\cite{baldrati2023searle}, comparing against the Qwen2.5-VL baseline and ReCALL~(Tab.~\ref{tab:more_benchmarks}).
EnhFashionIQ and Shoes are evaluated with the FashionIQ-trained models, and CIRCO is evaluated with the CIRR-trained models.

\noindent\textbf{enhFashionIQ.} \ours improves R@1 by $6.95$ over the baseline and $3.15$ over ReCALL, with consistent gains at R@10 and R@50.

\noindent\textbf{Shoes.} \ours also transfers to Shoes, improving R@1 by $2.38$ over the baseline and $1.13$ over ReCALL.
Reranking adds $+0.51$ R@1 and $+0.28$ R@10, while R@50 is best with the \eos embedding alone.

\noindent\textbf{CIRCO.} On CIRCO dataset, we observe that the \eos embedding of \ours achieves the best mAP at all cutoffs, improving mAP@5 by $0.64$ over the baseline and $0.73$ over ReCALL.
Reranking, however, does not help here: it lowers mAP@5 by $0.59$, back to the level of the baseline.

\noindent\textbf{Discussion.} Reranking remains beneficial when the target domain is close to the training domain~(enhFashionIQ, Shoes), but not under a larger domain shift~(CIRR $\rightarrow$ CIRCO).
We attribute this to the latents being trained to factorize targets of a single domain, which makes them less transferable than the \eos embedding.
We therefore recommend using the \eos embedding alone in such cases, which is also the cheaper option.
Whether training the latents jointly on diverse domains makes reranking generalize is left for future work.

\begin{table*}[t!]
    \centering
    \caption{\textbf{Sensitivity to the number of latents $K$.}
    We compare the Stage~2 models~(without post-training) between each configuration.
    Avg. on Refined-CIRR is the mean of R@1 and R$_{\mathrm{subset}}$@1.
    }
    \label{tab:num_latents}
    \setlength{\aboverulesep}{0pt}
    \setlength{\belowrulesep}{0pt}
    \renewcommand{\arraystretch}{1.15}
    \setlength{\tabcolsep}{4pt}
    \footnotesize
    \begin{tabular}{c cccc ccc c cc cc cc cc}
        \toprule
        \multirow{3}{*}{$K$}
        & \multicolumn{8}{c}{Refined-CIRR}
        & \multicolumn{8}{c}{Refined-FashionIQ} \\
        \cmidrule(lr){2-9} \cmidrule(lr){10-17}
        & \multicolumn{4}{c}{Recall@N}
        & \multicolumn{3}{c}{Recall$_{\mathrm{subset}}$@N}
        & \multirow{2}{*}{Avg.}
        & \multicolumn{2}{c}{Dress}
        & \multicolumn{2}{c}{Shirt}
        & \multicolumn{2}{c}{Top\&Tee}
        & \multicolumn{2}{c}{Avg.} \\
        \cmidrule(lr){2-5} \cmidrule(lr){6-8} \cmidrule(lr){10-11} \cmidrule(lr){12-13} \cmidrule(lr){14-15} \cmidrule(lr){16-17}
        & R@1 & R@5 & R@10 & R@50
        & R@1 & R@2 & R@3 &
        & R@10 & R@50
        & R@10 & R@50
        & R@10 & R@50
        & R@10 & R@50 \\
        \midrule
        4
            & 67.08 & 90.66 & 95.29 & 98.93 & 86.02 & 95.32 & 98.40 & 76.55
            & 63.87 & 82.58 & 70.69 & 86.19 & 73.13 & 87.57 & 69.23 & 85.45 \\
        8
            & 66.55	& 90.33 & 95.11	& 98.91	& 85.54	& 95.47	& 98.17	& 76.05
            & 64.77 & 82.52 & 71.38 & 86.71 & 73.18 & 87.79 & 69.78 & 85.67 \\
        16
            & 66.80 & 90.30 & 95.01 & 99.01 & 86.38 & 95.37 & 98.27 & 76.59
            & 64.94 & 83.53 & 70.38 & 86.82 & 73.13 & 88.00 & 69.48 & 86.12 \\
        \bottomrule
    \end{tabular}
\end{table*}
\section{Results with Stronger Backbones}
\label{sec:supp_backbone}
In the main paper, we applied \ours to stronger MLLM backbones.
In this section, we report the full results together with the baseline fine-tuned on
each backbone under the same setting~(Tab.~\ref{tab:backbone2}).
On Qwen3-VL-8B, \ours improves over the baseline on every metric, with
$+3.08$ R@1 on Refined-CIRR and $+4.21$ average R@10 on Refined-FashionIQ.
On Qwen3-VL-Embedding-8B, a backbone already specialized for retrieval, the gains are smaller but remain consistent on the top-ranked metrics: $+1.68$ R@1 on Refined-CIRR and $+1.90$ average R@10 on Refined-FashionIQ.
We attribute the slight decrease in R@5 to R@50 on Refined-CIRR to saturation: both models exceed $93$ on these metrics and differ by at most $0.31$.

\section{Effect of Auxiliary Objectives on Efficient Reranking }
\label{sec:supp_rerank}
As mentioned in Sec.~4.3 in the main paper, we argued that the gain of efficient reranking~(ERR) does not stem from merely fusing an additional vector per image.
In Tab.~\ref{tab:rerank_control}, we validate this claim by comparing the impact of our reranking process on top of the na\"ive baseline with additional latents inserted before \eos~(b) and our Stage 2 model~(d).
Note that the only difference between the two is the use of our auxiliary objectives~($\mathcal{L}_{\mathrm{imag}}$, $\mathcal{L}_{\mathrm{distill}}$, and the factor-level InfoNCE).
We then apply the identical reranking rule~(Eq.~9).
To illustrate, while ERR clearly improves our model~(e), it is shown that ERR rather degrades the performance of the baseline on the FashionIQ dataset when the latents are not given any role.
These results indicate that the gain cannot be attributed to the extra vector alone; the mean factor becomes complementary to [EOS] only when the latents are shaped by our objectives.

\section{Sensitivity to the Number of Latents}
We vary the number of latents $K \in \{4, 8, 16\}$ while keeping all other settings fixed~(Tab.~\ref{tab:num_latents}).
Performance is stable across $K$: varying the number of latents shifts Avg. on Refined-CIRR by at most 0.54 and Avg. R@10 on Refined-FashionIQ by at most 0.55, and no value of $K$ is consistently best across the two benchmarks.
This suggests that \ours does not rely on a carefully chosen number of latents within the tested range~($K \geq 4$). 
We conjecture that a small number of factors already suffices to cover the regions typically affected by an edit in these benchmarks.
We use $K{=}8$ as the default in all experiments.

\section{Qualitative Results}
\begin{figure*}[t]
    \centering
    \vspace{-0.2cm}
    \includegraphics[width=1\textwidth]{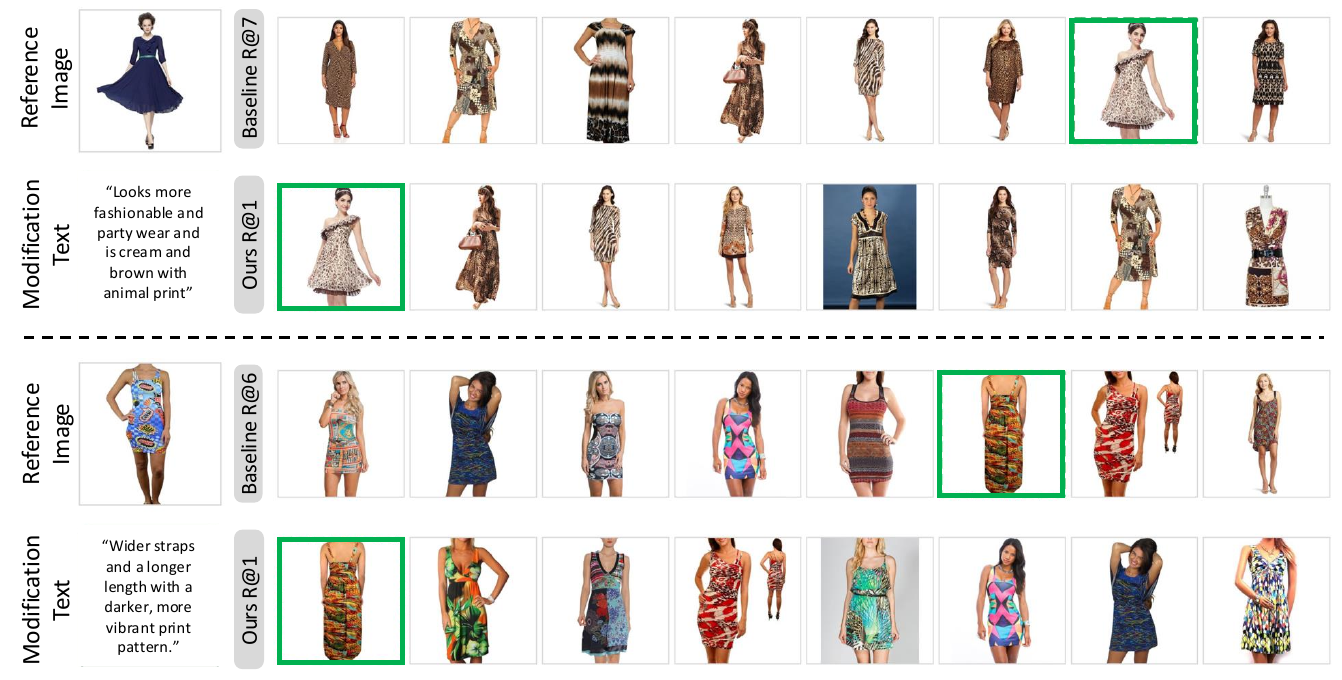}
    \vspace{-0.1cm}
    \caption{
        \textbf{Results on dress category.}
    }
    \label{fig:fiq_qual_dress}
\end{figure*}

\begin{figure*}[t]
    \centering
    \vspace{-0.2cm}
    \includegraphics[width=\linewidth]{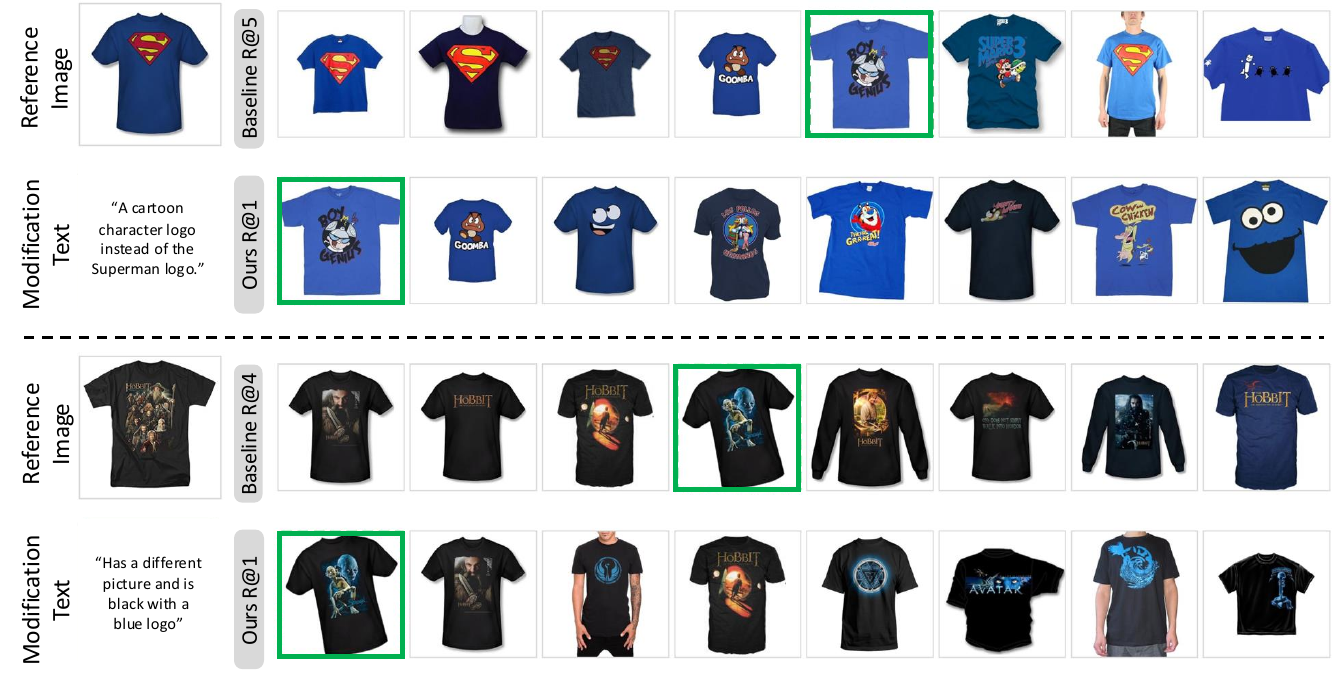}
    \vspace{-0.1cm}
    \caption{
        \textbf{Results on Shirt category.}
    }
    \label{fig:fiq_qual_shirt}
\end{figure*}

\begin{figure*}[t]
    \centering
    \vspace{-0.5cm}
    \includegraphics[width=\linewidth]{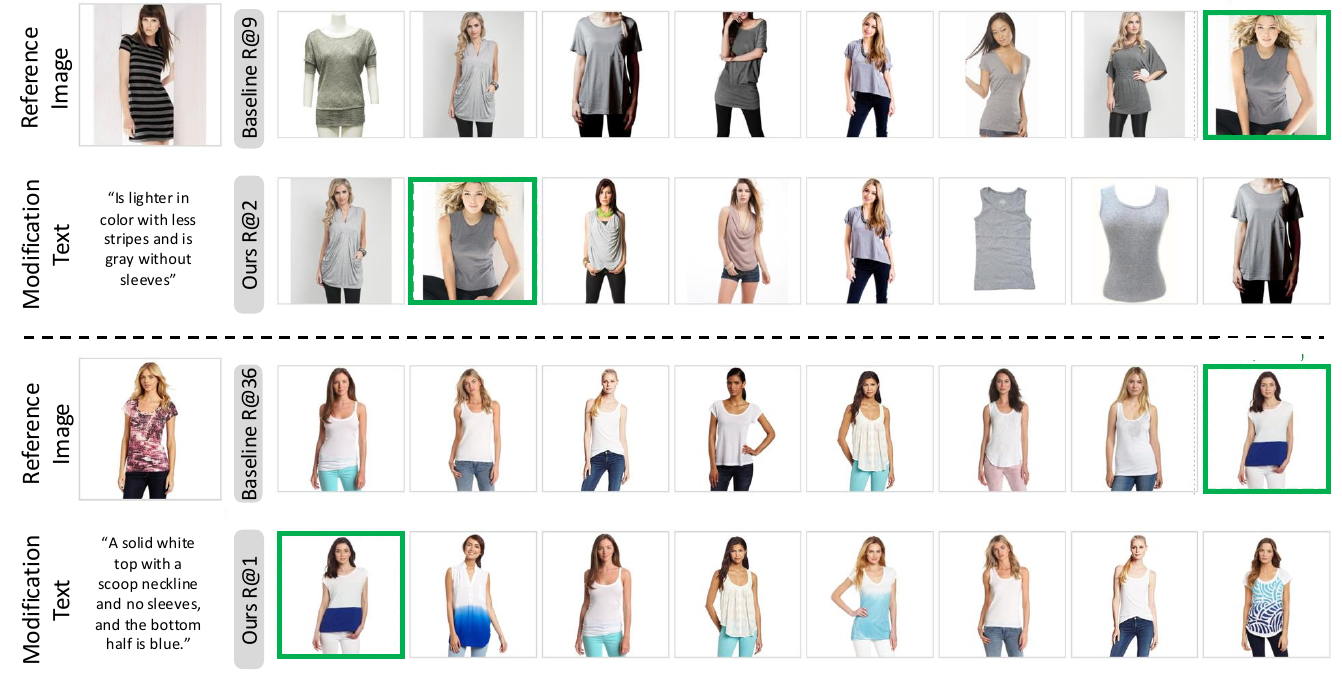}
    \vspace{-20pt}
    \caption{
        \textbf{Results on Top\&Tee category.}
    }
    \label{fig:fiq_qual_toptee}
\end{figure*}

\begin{figure*}[t]
    \centering
    \vspace{-5pt}
    \includegraphics[width=1.\textwidth]{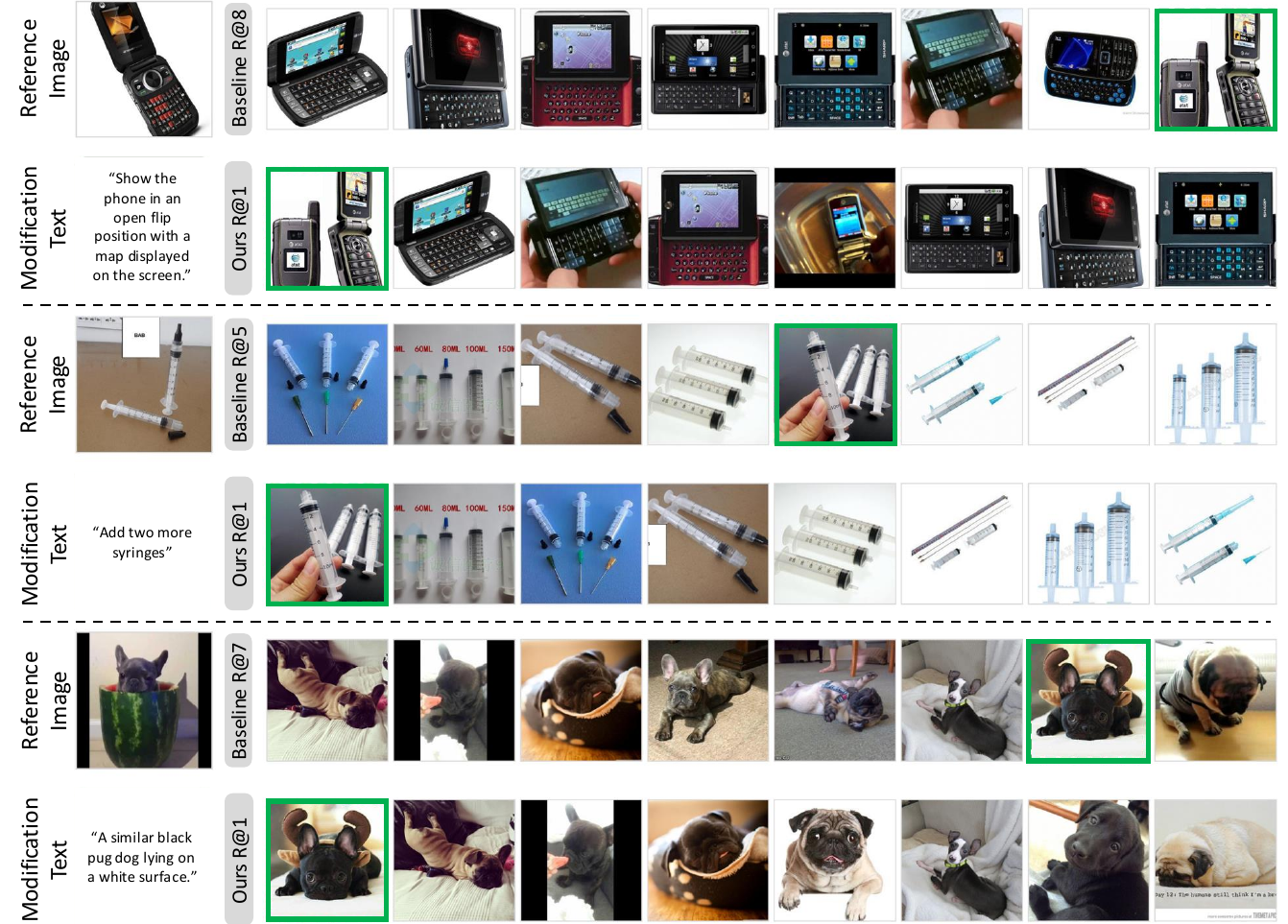}
    \vspace{-15pt}
    \caption{
        \textbf{Results on CIRR.}
    }
    \label{fig:qual_cirr}
\end{figure*}

In Fig.~\ref{fig:fiq_qual_dress}, Fig.~\ref{fig:fiq_qual_shirt}, Fig.~\ref{fig:fiq_qual_toptee}, and Fig.~\ref{fig:qual_cirr}, we present qualitative comparisons between \ours and our baseline~(Qwen2.5-VL-7B-Instruct~\citep{bai2025qwen25vl} fine-tuned via DoRA~\citep{liu2024dora}).
In CIR, modifying a reference image with text often projects the query into a broad semantic space where multiple target images can be considered valid matches. 
Therefore, rather than restricting our analysis to Top-1 prediction, we examine the Top-8 retrieved results to better assess the overall alignment between the composed query and the retrieved distribution.

In the first example of Fig.~\ref{fig:fiq_qual_dress}, which belongs to the dress category of FashionIQ dataset, the query specifies an item that is ``more fashionable and party wear and is cream and brown with animal print''.
While both models successfully capture the desired color scheme, \ours ranks candidates that accurately reflect the fine-grained ``animal print'' pattern higher than the baseline. 
In the second example, the baseline struggles with structural modifications, incorrectly retrieving an item with short sleeves at Rank-2 and entirely ignoring the requested ``strap'' detail at Rank-3. 
In contrast, the top-ranked candidates of \ours better reflect these structural constraints.
Fig.~\ref{fig:fiq_qual_shirt} further highlights a common failure mode in standard contrastive learning, which fails to handle negative constraints. 
For the query containing the phrase ``instead of superman logo'', the baseline fails to disentangle the negated element, retrieving images containing the superman logo. 
In contrast, \ours successfully suppresses the negated visual element while maintaining the core semantics of the reference image.

Similar trends are observed on open-domain CIRR dataset. 
As shown in the first example of Fig.~\ref{fig:qual_cirr}, the query requires the model to identify ``map displayed on the screen'' of a smartphone. 
While the baseline is easily distracted by generic smartphone images, \ours retrieves the target showing a map on the screen at Rank 1.
Also, for a query involving relative counting~(second example), the baseline ranks the target lower than FactEmbed (Rank 5 vs. Rank 1).
Overall, we find that the baseline trained with standard contrastive learning often ignores the comparative conditions, while \ours captures these relationships more faithfully.


\clearpage\clearpage

\bibliography{main}
\bibliographystyle{unsrt}
\end{document}